\documentclass[11pt,a4paper]{article}

\usepackage{liatuo}          

\title{MachEmbodied-U0: Unified Understanding and Generation Model for Embodied Intelligence}
\author{\textbf{Foundation Model, Li Auto Inc.}}
\shorttitle{ME-0 Technical Report}
\reportdate{September 2026}

\begin{document}

\makeLiTitle

\begin{liabstract}
General-purpose robot control requires models to understand task intent,
identify where to interact, capture how the scene evolves, and generate
precise actions. Vision-language-action models provide strong semantic
priors but typically do not explicitly model scene dynamics, while
world-action models couple visual prediction with control without
necessarily exposing the task-relevant semantic and spatial structure
needed for fine-grained manipulation.
We present \textbf{MachEmbodied-U0 (ME-U0)}, a unified embodied foundation model connecting understanding and generation experts through a Mixture-of-Transformers architecture. Subtask prediction and affordance grounding guide joint visual-dynamics and action generation via flow matching. Visual dynamics encompass future RGB, depth, surface normals, and optical flow, providing complementary supervision for appearance, geometry, and motion. Multi-rate Rotary Position Encoding (MRPE) aligns visual dynamics with fine-grained control.
We pretrain ME-U0 on approximately 4,200 hours of curated demonstrations from robotic datasets and egocentric datasets. Using only the
supervision natively available in each downstream benchmark, ME-U0
achieves an average score of 17.66 on the RoboDojo simulation benchmark
and average success rates of 99.0\% and 82.5\% on LIBERO and LIBERO-Plus,
respectively. We additionally validate ME-U0 on real-world robotic
manipulation tasks, demonstrating its effectiveness beyond simulation.
Without corresponding downstream supervision, ME-U0 further demonstrates zero-shot subtask prediction, affordance grounding, and visual dynamics on simulated and real-world observations. Overall, ME-U0 combines competitive downstream control
performance with transferable task-grounding and visual-dynamics
capabilities across simulation and the real world.

\vspace{1em}
\noindent\textit{\textbf{Project Page:}\enspace\url{https://machembodied.com/ME-U/ME-U0.html}}

\noindent\textit{\textbf{Github:}\enspace\url{https://github.com/MachEmbodied/ME-U0}}
\end{liabstract}

\newpage
\tableofcontents


\begin{figure}[t]
    \centering
    \includegraphics[width=\linewidth]{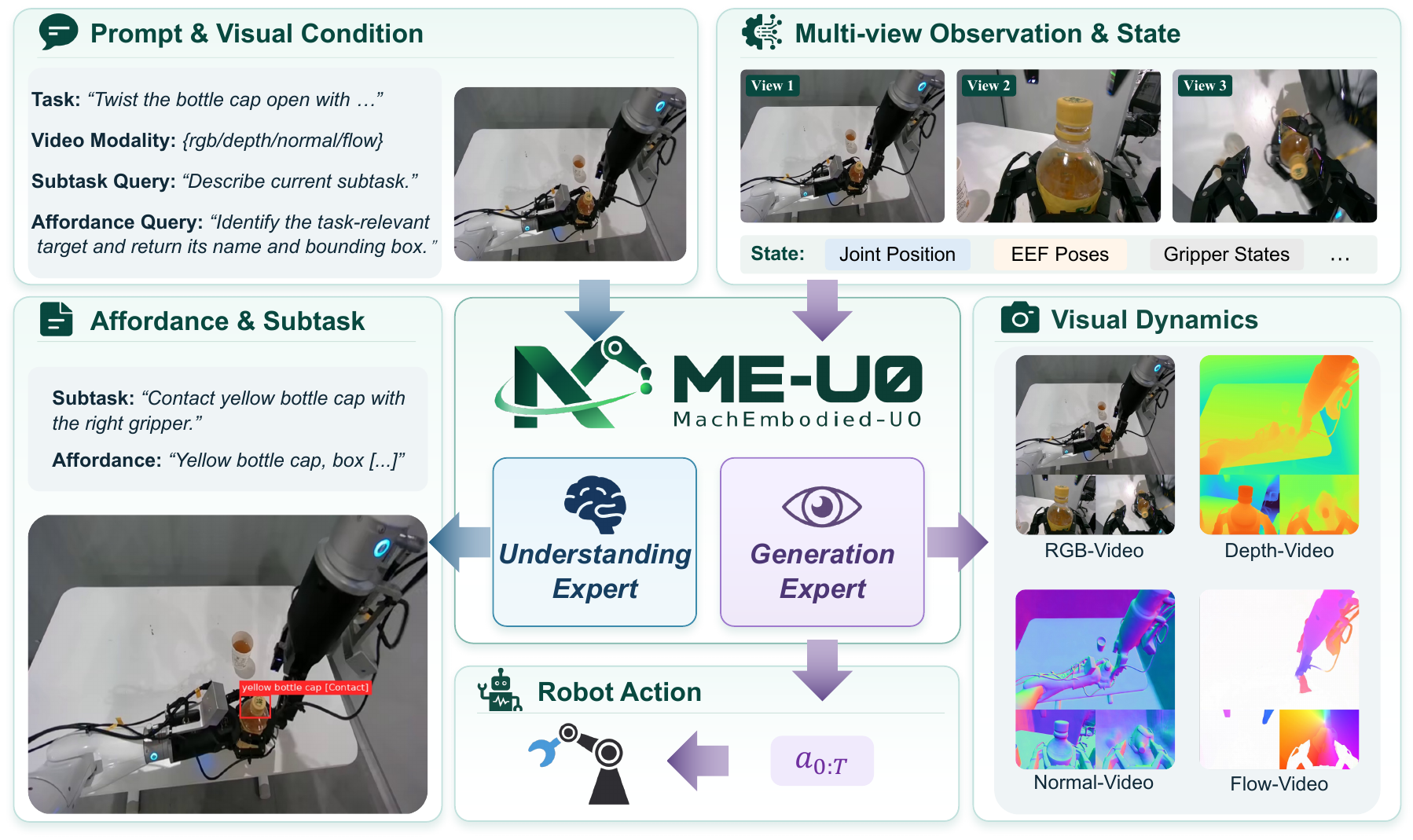}
    \caption{MachEmbodied-U0 (ME-U0) integrates task-grounded
    understanding with joint visual and action generation through
    cooperating understanding and generation experts.}
    \label{fig:teaser}
\end{figure}

\section{Introduction}
\label{sec:introduction}

Physical intelligence requires more than recognizing objects or following
instructions: a robot must determine what to do, where to interact, and
how its actions will change the world. These requirements connect semantic
understanding, prediction, and control. A general embodied foundation
model should therefore ground task intent in the observed scene while
relating executable actions to their anticipated physical consequences.

Vision-language-action (VLA) models leverage the semantic priors of
pretrained vision-language models for language-conditioned robot
control~\cite{blip2,llava,bai2025qwen25vltechnicalreport,rt2,openvla,octo,pi0,pi05,fast}.
However, their action-centric objectives typically do not explicitly
capture how the scene evolves during interaction. World-action models
(WAMs) address this limitation by jointly modeling future observations
and robot actions, providing dense supervision for spatial structure and
motion~\cite{unipi,gr1,dreamzero,lawam,fastwam}. Yet visual dynamics alone
does not ensure task-relevant grounding, interaction localization, or
long-horizon task decomposition. These complementary strengths motivate
a model that connects semantic understanding and visual dynamics with
continuous control.

Unified multimodal architectures offer a natural basis for this
integration~\cite{showo,bagel,lance}. Recent embodied
extensions explore several forms of unification: 
UAM ~\cite{uam} separates semantic and visuomotor pathways to preserve multimodal
competence during action learning; Motubrain ~\cite{motubrain} jointly models
language-conditioned video and actions across policy, world-modeling,
and inverse-dynamics modes; and Cosmos~3 ~\cite{cosmos3} scales unified reasoning and
generation across language, vision, audio, and actions.

Together, these advances show that semantic reasoning, visual dynamics,
and control can coexist within a unified architecture. They also expose
a remaining design question: interleaving outputs, separating expert
pathways, or supporting flexible modality configurations does not by
itself determine how task-level intent should be translated into
spatially grounded and dynamically informed control. For manipulation,
the central challenge is therefore not merely to generate text, visual
observations, and actions, but to organize their interaction around four
coupled questions: what operation should be performed next, where the
interaction should occur, what visual, geometric, and motion changes
should unfold during execution, and how the operation should be realized
through continuous control. This motivates a manipulation-centered form
of unification that connects task decomposition and affordance grounding
with geometry- and motion-aware visual dynamics and action generation.

We introduce \textbf{MachEmbodied-U0 (ME-U0)}, a unified embodied foundation model capable of subtask decomposition, scene understanding, visual dynamics generation, and continuous action generation. As illustrated in Fig.~\ref{fig:teaser}, the model
couples an understanding expert with a generation expert in a
Mixture-of-Transformers (MoT) architecture. Conditioned on task
instructions, visual observations, and robot states, the two experts
connect scene-grounded task interpretation with visual dynamics generation
and continuous control. Specialized parameters accommodate the distinct
demands of semantic understanding and continuous generation, while shared
multimodal attention enables coordination between them. ME-U0 combines three complementary
designs: explicit subtask reasoning and affordance grounding, joint
multimodal visual-dynamics and action generation, and action-substep
temporal alignment. Together, these designs establish a direct pathway
from high-level intent and local interaction geometry to anticipated
scene evolution and executable control.

The understanding expert predicts subtasks and affordances to specify
\emph{what to do next} and \emph{where to act}. Subtask prediction
connects the overall instruction to the current manipulation stage,
while affordance prediction identifies and localizes the relevant
interaction target in the observed scene. Together, these outputs
translate high-level intent into actionable semantic context.
Through shared multimodal attention, the generation expert can use
this context to condition future visual and action predictions,
integrating task understanding into the control process.

Conditioned on this semantic context, the generation expert jointly
predicts visual dynamics and continuous actions through
Flow Matching~\cite{flow_matching}, coupling control with anticipated
scene changes in a shared backbone. We extend RGB generation to depth,
surface normals, and optical flow~\cite{genception}, adding explicit
geometry and motion supervision to support precise manipulation.
MRPE aligns action steps with
temporally compressed visual transitions.

To scale pretraining across heterogeneous robot embodiments, we map
embodiment-specific states and actions into a canonical interface while
preserving their physical control semantics. Dataset-specific
transformations align joint, gripper, base, torso, and head variables,
while validity masks indicate the dimensions and time steps available
for each sample. This interface enables a single model to learn from
approximately 4,200 hours
of curated robotic and egocentric demonstrations, with the robotic
corpus spanning four datasets and six embodiments. Through this
pretraining, ME-U0 jointly acquires task-grounded understanding,
geometry- and motion-aware visual dynamics, and continuous control.

For downstream evaluation, we use only the supervision natively
available in each benchmark, without introducing additional subtask,
affordance, geometry, or motion annotations during post-training.
ME-U0 achieves a score of 17.66 on RoboDojo and average success rates
of 99.0\% and 82.5\% on LIBERO and LIBERO-Plus, respectively. These
results demonstrate that the representations acquired during
large-scale pretraining transfer effectively to downstream manipulation,
even under reduced-supervision post-training.

Our contributions are fourfold:
\begin{itemize}
    \item \textbf{A unified embodied foundation model.}
    We present ME-U0, which integrates task-grounded understanding,
    geometry- and motion-aware visual dynamics, and continuous robot
    control within a shared Mixture-of-Transformers architecture.

    \item \textbf{Grounded visual--action dynamics modeling.}
    Subtask prediction and affordance grounding provide structured
    semantic and spatial context for generation. Future RGB, depth,
    surface normals, optical flow, and continuous actions are jointly
    modeled through flow matching, while MRPE aligns
    fine-grained control with temporally compressed visual dynamics.

    \item \textbf{A scalable embodied pretraining system.}
    We construct a heterogeneous pretraining mixture with unified robot
    state--action semantics, multimodal annotations, task-aware sampling,
    and efficient data infrastructure, enabling training across four
    datasets and six robot embodiments.

    \item \textbf{Broad empirical validation.}
    ME-U0 achieves strong downstream performance on RoboDojo, LIBERO,
    and LIBERO-Plus, and further demonstrates effective transfer to
    real-world robotic manipulation.
\end{itemize}

\section{Related Work}

\subsection{Vision-Language-Action Models}

Vision-language-action (VLA) models ground visual observations and language instructions in robot actions. RT-1 established the benefits of large-scale,
multi-task robot learning, while RT-2 incorporated pretrained vision-language
knowledge into control through action-token prediction~\cite{rt1,rt2}.
Open generalist policies further improve accessibility and adaptation:
Octo accommodates new observation and action spaces, and OpenVLA supports
efficient fine-tuning of a pretrained vision-language backbone for
manipulation~\cite{octo,openvla}.

Another line of work focuses on action representation and generalization.
FAST compresses action sequences using frequency-domain tokenization, whereas
$\pi_0$ uses flow matching to generate continuous actions, offering different
approaches to modeling dexterous control~\cite{fast,pi0}. Building on $\pi_0$,
$\pi_{0.5}$ combines data from multiple robots and the web with high-level
semantic prediction, enabling long-horizon manipulation in previously unseen
homes~\cite{pi05}. Together, these methods highlight the complementary roles of
transferable semantic knowledge, diverse training data, and expressive action
representations in generalist robot policies.

\subsection{World Action Models}

World action models (WAMs) connect visual dynamics modeling with action
learning, using future prediction to inform control or shape policy
representations. Early approaches include UniPi, which extracts actions from
language-conditioned video plans, and GR-1, which jointly predicts future
images and robot actions after video generative pre-training~\cite{unipi,gr1}.
DreamZero scales joint video--action modeling with a pretrained video diffusion
backbone, supporting zero-shot policy generalization and transfer from
video-only demonstrations across embodiments~\cite{dreamzero}.

Recent methods examine how much future generation is necessary for effective
control. LaWAM predicts compact latent visual subgoals to condition action
generation, reducing the cost of reconstructing future video. Fast-WAM retains
video co-training but omits future prediction at inference, showing that
predictive supervision can benefit policies without requiring explicit
test-time imagination~\cite{lawam,fastwam}. These approaches expose an important
design choice: whether to use predicted futures as inference-time context or
primarily as supervision for learning representations of scene dynamics.

\subsection{Unified Models}

Unified multimodal models combine understanding and generation through
shared multimodal context. Show-o integrates autoregressive language modeling
with discrete diffusion for visual generation, while BAGEL scales unified
pre-training on interleaved multimodal data~\cite{showo,bagel}. Lance further
explores shared context with separate expert pathways for image and video
understanding, generation, and editing~\cite{lance}. These models provide
foundations for coupling semantic reasoning with visual prediction.

For embodied control, BagelVLA interleaves language planning, visual forecasting,
and action generation to support long-horizon manipulation, while Cosmos~3
extends a unified mixture-of-transformers architecture to language, image,
video, audio, and action sequences~\cite{bagelvla,cosmos3}. UAM addresses
multimodal forgetting during action training by introducing a parallel Dorsal
Expert initialized from a generative model and supervised to predict visual
dynamics~\cite{uam}. Its separation of semantic and control-related processing
complements joint generation approaches, emphasizing that integrating
perception, prediction, and action also requires preserving pretrained
understanding capabilities.

\section{Training Data}
\label{sec:data}
We draw on a raw pool of approximately 5,700 hours of robotic demonstrations and 3,920 hours of egocentric demonstrations.
Egocentric data complements robotic data by covering manipulation contexts and behaviors underrepresented in robot demonstrations.
After careful curation, we retain approximately 4,200 hours of demonstrations for pretraining.
We further annotate a subset of the data with subtask, affordance, geometry, and motion labels to support joint understanding and generation.

\subsection{Dataset Composition}
\subsubsection{Robotic Datasets} 

\begin{figure*}[t]
    \centering
    \includegraphics[width=\linewidth]{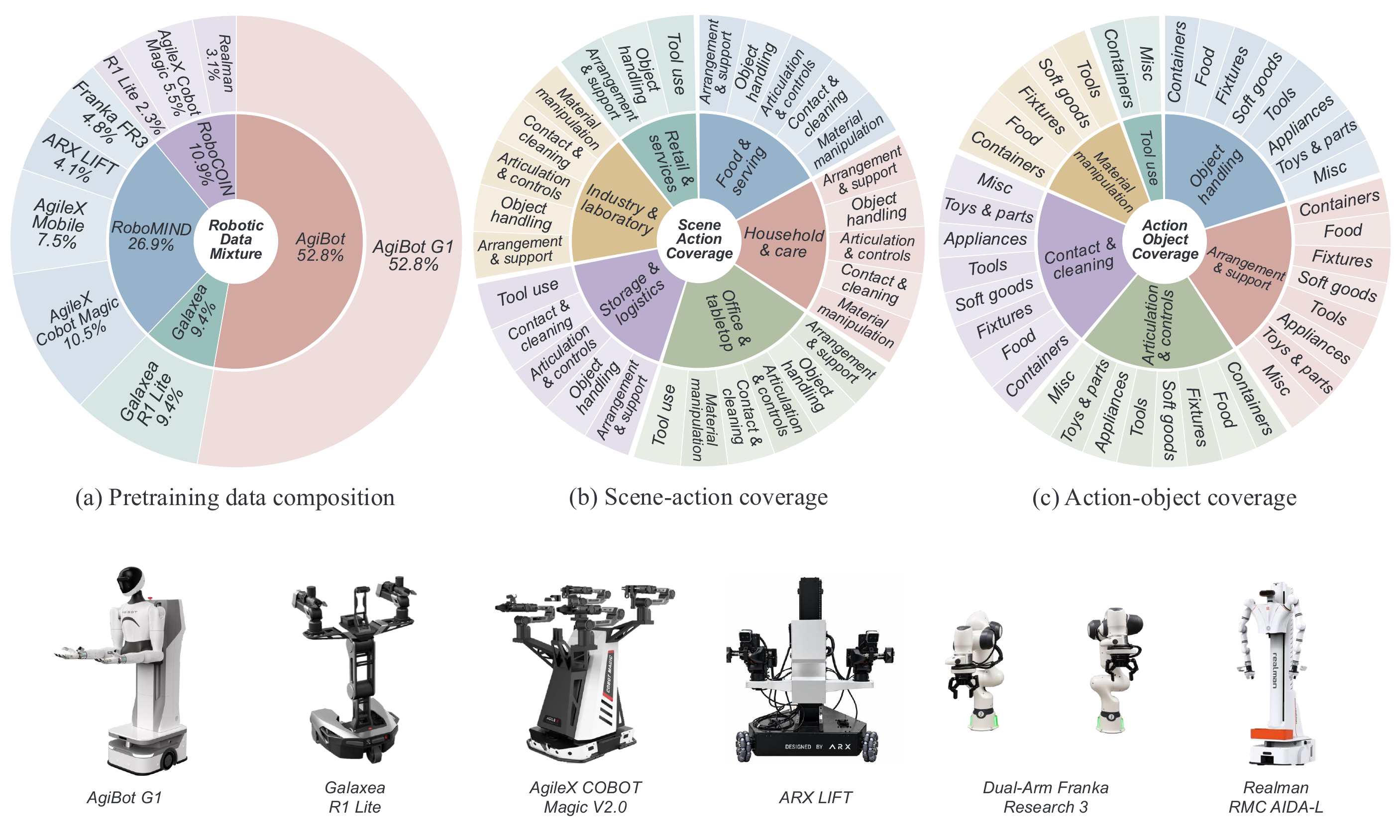}
    \caption{
    \textbf{Robotic data composition and \emph{scene--action--object} diversity.}
    (a) Dataset and robot-embodiment composition.
    (b) Task descriptions and the actions in each task.
    (c) Actions and the objects in task descriptions.
    }
    \label{fig:pretraining-data-overview}
\end{figure*}

Our robotic data comes from four open-source datasets:
AgiBot World Beta~\cite{bu2025agibot},
RoboMIND 2.0~\cite{hou2026robomind20multimodalbimanual},
RoboCOIN~\cite{wu2026robocoinopensourcedbimanualrobotic},
and Galaxea Open-World~\cite{galaxea2025}.
After validity checks, the resulting corpus spans six robot embodiments and a broad range of manipulation tasks.
Fig.~\ref{fig:pretraining-data-overview}(a) summarizes its dataset and embodiment composition.

Beyond data sources and embodiments, we construct a \emph{scene--action--object} tag scheme. The three dimensions describe the scene context of a task, the behaviors it involves, and the objects it mentions.
For visualization clarity, we aggregate related tags into six scene groups, six action groups, and eight
object groups. Fig.~\ref{fig:pretraining-data-overview}(b) summarizes scene--action coverage. Common behaviors recur across different task contexts, while each context encompasses multiple forms of manipulation. This overlap places shared skills in varied settings, extending the mixture beyond a narrow set of context-specific behaviors. Fig.~\ref{fig:pretraining-data-overview}(c) complements this view with action--object coverage, showing how shared behaviors are associated with objects of different physical properties and functional roles.

The data mixture connects a broad range of task meanings with concrete robot interactions across diverse scenes and objects, providing rich supervision for grounding semantic understanding in physical behavior and supporting transfer across tasks.

\subsubsection{Egocentric Datasets}
To improve coverage of task tags underrepresented in the robotic datasets, 
we select egocentric demonstrations with
matching tags and incorporate them into the overall pretraining mixture. For action-supervised pretraining, we primarily use EgoDex~\cite{hoque2026egodex}, EgoLive~\cite{li2026egolive,zhang2026joyai}, EgoVerse~\cite{punamiya2026egoverse}, HOI4D~\cite{liu2022hoi4d}, HOT3D~\cite{banerjee2025hot3d}, and the hand-annotated portion of Ego-Exo-4D~\cite{grauman2024ego}. We select demonstrations with temporally aligned first-person observations and low-level human motion annotations, including fingertip locations, wrist poses, and arm or shoulder states. These annotations support conversion into the unified hand-centric action representation described in Section~\ref{sec:ego_data_processing}.

\subsection{Robotic Data Processing}
\subsubsection{Data Curation}

\begin{figure}[t]
    \centering
    \includegraphics[width=\linewidth]{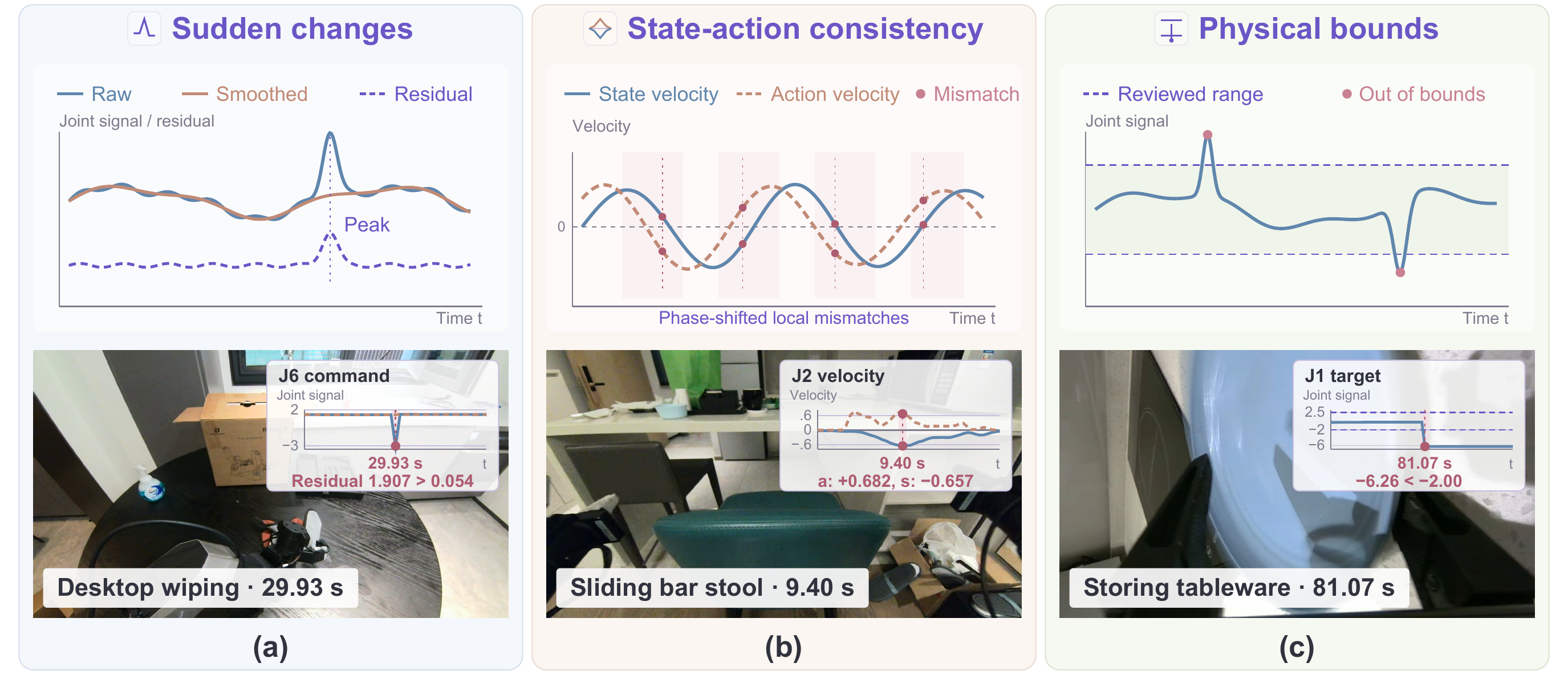}
    \caption{Schematic of real-robot data curation. Each stage removes a subset of the episodes passed to it.}
    \label{fig:real_robot_data_filtering}
\end{figure}

Fig.~\ref{fig:real_robot_data_filtering} illustrates our three-stage cleaning pipeline and representative signal anomalies. Following recent robot-data curation pipelines~\cite{wu2026foundation,yuan2026qwen},
we adopt dataset-specific filtering policies to account for differences in state definitions, units, sampling rates, and action conventions. Before filtering, source-specific readers establish the physical meaning, units, ordering, and temporal alignment of the recorded channels. Each source is evaluated using checks supported by its available signals.

We apply three checks in sequence.
First, we detect sudden changes in continuous state and action channels after scaling each dimension by its $q_{99}-q_{01}$ range. A median filter followed by smoothing provides a reference trajectory; large residuals, accelerations, or jerks flag an episode. Second, we assess state--action consistency by estimating the command--response lag from cross-correlations of smoothed first differences, then measuring directional agreement at lag-aligned timesteps where actions vary. Low agreement across the configured number of dimensions, or a state freeze during action variation,
flags the episode. Third, we compare raw values with manually reviewed, per-dimension physical bounds informed by full-dataset statistics. Gripper channels are excluded from these continuous-signal checks because their open/close values do not represent smooth motion. An episode is excluded if any enabled check is triggered or a source-specific timestamp-gap rule is violated. Filtering operates on complete episodes and leaves the original recordings unchanged, preserving the temporal continuity of retained demonstrations.

\subsubsection{Data Annotation}

We augment selected portions of the data with task-grounded semantic, spatial, geometric, and motion annotations.

\begin{figure*}[t]
    \centering
    \makebox[\textwidth][c]{%
        \includegraphics[width=0.99\textwidth]{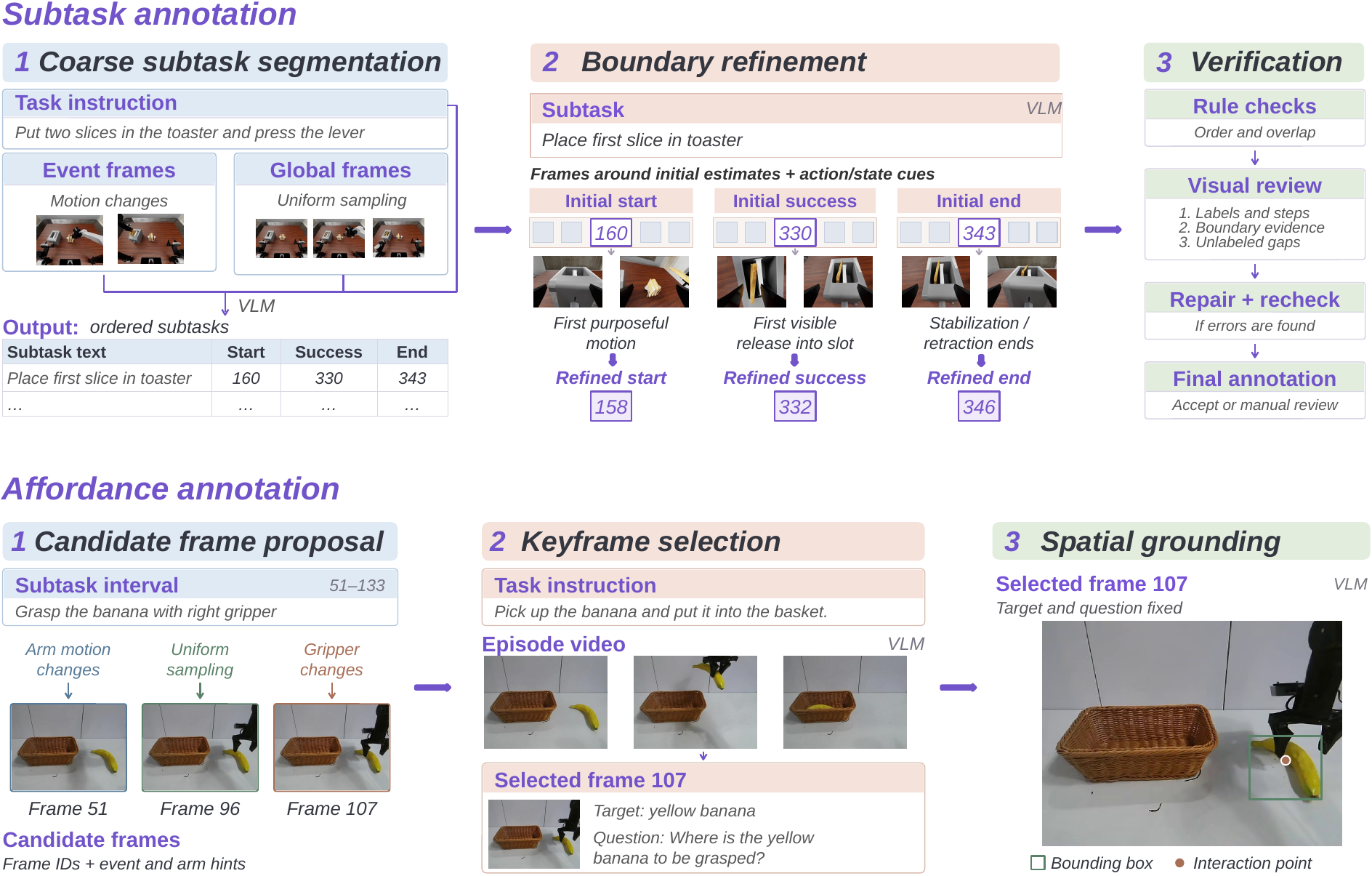}%
    }
    \par\vspace{6pt}
    \caption{
        Overview of our data annotation pipeline.
        We annotate temporal subtasks through segmentation, boundary
        refinement, and verification, and obtain spatial affordances
        through keyframe selection and grounding.
    }
    \label{fig:data_annotation}
\end{figure*}

\noindent\textbf{Subtask Labeling.}  
We develop a three-stage pipeline for labeling robot demonstrations with temporally grounded subtasks shown in Fig.~\ref{fig:data_annotation}. Our annotation pipeline comprises three stages. 
\textit{Coarse subtask segmentation} combines event timestamps identified from action and state changes with uniformly sampled timestamps, capturing both salient transitions and the overall task progression. Using synchronized multi-view observations at these timestamps and the task instruction, the pipeline identifies an ordered sequence of subtasks and their approximate temporal intervals.
\textit{Boundary refinement} then examines visual observations near the proposed timestamps, together with temporally aligned action and state changes, to refine each subtask's start and end boundaries and identify the first frame where its goal is visibly achieved.
\textit{Verification} checks the resulting sequence for semantic coherence, temporal ordering, and boundary consistency. Annotations that fail verification receive one automatic revision, with unresolved cases referred for manual review. We use GPT-5.5~\cite{openai2026gpt55} for all stages.

\noindent\textbf{Affordance Labeling.} Affordance annotations identify the target object and interaction location at each manipulation stage, thereby bridging high-level task semantics and low-level robot actions. As illustrated in Fig.~\ref{fig:data_annotation}, the annotation pipeline comprises three stages.
\textit{Candidate frame proposal} uses a rule-based procedure adapted from AffordanceVLA~\cite{yu2026affordancevla} to extract candidate frames from each demonstration.
\textit{Keyframe selection} jointly examines the head-camera video, task instruction, subtask timeline, and candidate contact sheet. It selects frames depicting meaningful interactions and assigns each a target category and an affordance instruction.
The \textit{Spatial grounding} stage localizes the target object and its interaction point in each selected frame. Frames are upsampled to twice their original resolution for finer localization; predicted coordinates are then mapped back to the original image space and checked for spatial consistency. 
The model we use is Qwen3.6-35B-A3B~\cite{qwen36_35b_a3b}.

\textbf{Visual Dynamics Labeling.}
We derive depth, surface normals, and optical flow labels from the original RGB data
to provide complementary geometric and motion supervision for our unified model.
These auxiliary annotations are intended to strengthen the model's spatial and
temporal understanding, supporting its ability to reason about scene structure and
motion in robotic manipulation.

 We use MoGe-2~\cite{moge2} to estimate metric depth and surface normals
from individual RGB frames, and normalize the depth labels~\cite{genception}. To capture motion across frames, we use
WAFT~\cite{waft} to estimate optical flow between sampled frame pairs. We adjust the
temporal stride for each dataset according to its sampling frequency, since the same
frame offset can correspond to different time intervals across datasets. Together,
these annotations provide supervision for both scene geometry within individual
frames and visual motion across time.

\subsection{Egocentric Data Processing}
\label{sec:ego_data_processing}

\begin{figure}[t]
    \centering
    \includegraphics[width=\linewidth]{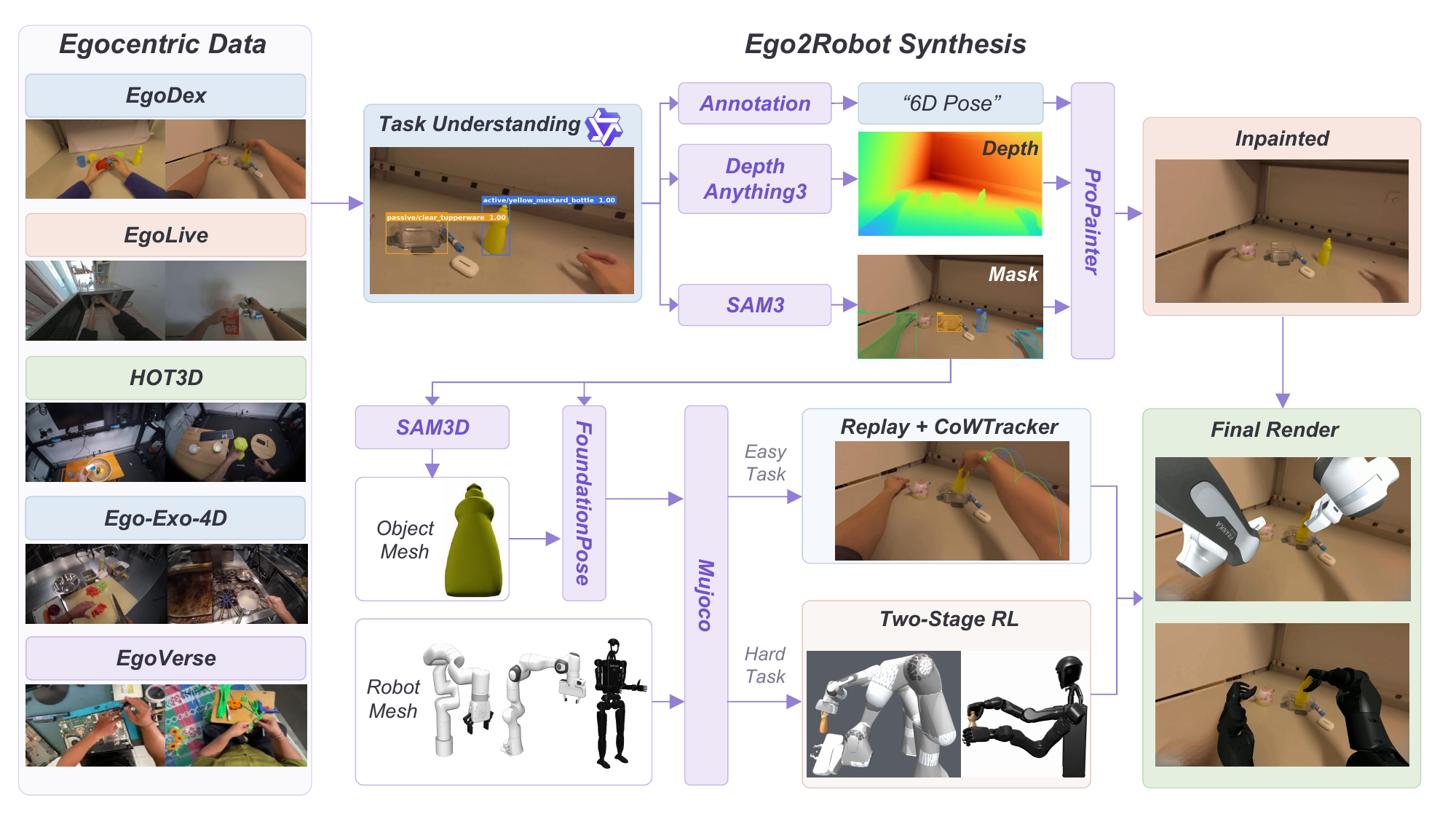}
    \caption{Overview of egocentric data processing. Ego-to-Robot synthesis converts human demonstrations into robot-aligned observations and trajectory annotations.}
    \label{fig:ego-processing}
\end{figure}

\noindent\textbf{Annotation Standardization.} Action-annotated egocentric demonstrations are standardized into a unified human-centric action format following LeRobot v3~\cite{cadene2024lerobot}, with invalid frames and discontinuous action chunks filtered before task-level normalization. Each action-supervised sample is centered on an anchor timestep and contains first-person observations, future action chunks, available language supervision, and hand states represented in end-effector space. Hand states are represented by 3D position and axis-angle orientation, providing a consistent state-action space across heterogeneous human trajectories.

\noindent\textbf{Ego2Robot Synthesis.}
As illustrated in Fig.~\ref{fig:ego-processing}, we synthesize robot-aligned demonstrations through task understanding, scene reconstruction, trajectory optimization, and rendering to reduce embodiment mismatch and depth uncertainty while improving contact accuracy.

We first use Qwen3.6~\cite{qwen36_35b_a3b} to parse each egocentric video and its instruction, identifying relevant objects, interaction roles, and physical properties. Based on these attributes, each episode is categorized as either an easy task suited to trajectory replay or a hard task requiring contact refinement. For both task categories, the scene is reconstructed using metric depth from Depth Anything 3~\cite{lin2025depth} and masks from SAM3~\cite{carion2026sam}. Following Inpaint Anything~\cite{yu2023inpaint}, we then apply ProPainter~\cite{zhou2023propainter} to the arm and hand masks to remove human regions and recover the occluded background for all episodes. 

Trajectory processing differs between the two task categories. For easy tasks, we replay the annotated trajectories and use CoWTracker~\cite{lai2026a} to ensure their accuracy in pixel space, thereby reducing the 2D-3D mismatch. For hard tasks involving contact sensitive interactions, we extend prior work~\cite{han2026video2sim2real,liu2026egoengine,wang2026ego2robot,yang2026handedit} to recover object geometry and motion. Manipulated rigid objects are reconstructed as 3D physical models by using SAM3D~\cite{sam3dteam2025sam3d3dfyimages}, while masks, depth, and 6D pose estimates~\cite{wang2025vggt,foundationposewen2024} are fused to recover temporally consistent object trajectories. The corresponding wrist and fingertip trajectories are then refined through a two-stage Proximal Policy Optimization (PPO)~\cite{schulman2017proximal} curriculum to improve contact accuracy and trajectory smoothness. The optimal trajectories from both branches are subsequently retargeted to the embodiment equipped with grippers through robot-specific inverse kinematics, solved with Mink~\cite{pan2025spiderscalablephysicsinformeddexterous} in MuJoCo~\cite{todorov2012mujoco}.
Finally, the robot mesh is rendered into the inpainted egocentric video, with occlusions resolved according to object depth. The rendered images and optimized trajectory annotations constitute the final robot-aligned dataset.

\section{Model Design}

\label{sec:model}
\subsection{Main Architecture}

\begin{figure}[t]
    \centering
    \includegraphics[width=\linewidth]
   {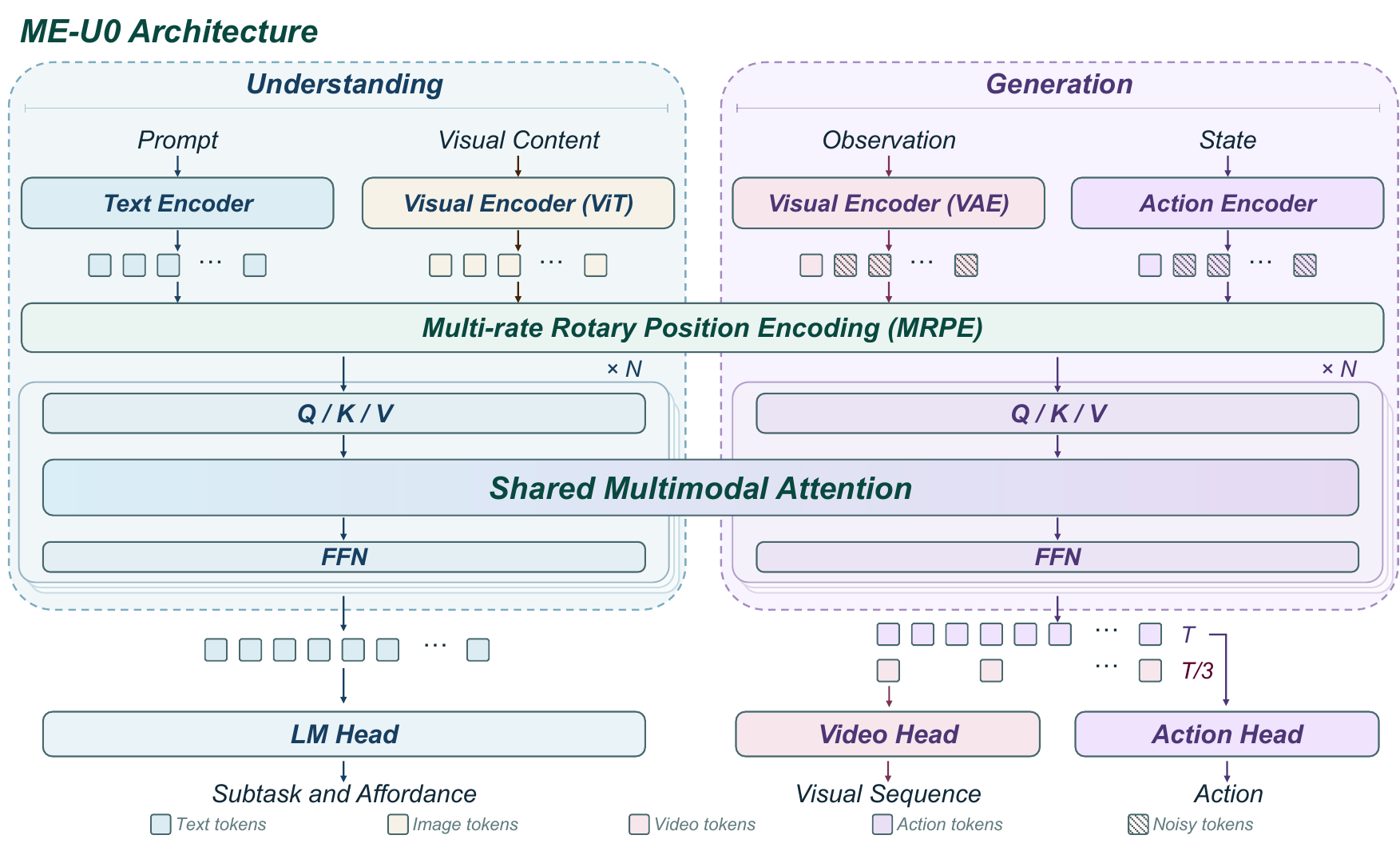}
    \caption{Overview of our framework. The understanding expert predicts subtasks and affordances, while the generation expert jointly predicts future visual observations and robot actions through shared multimodal attention.}
    \label{fig:pipeline}
\end{figure}
MachEmbodied-U0 integrates an understanding expert and a generation expert within a unified multimodal framework, as illustrated in Fig.~\ref{fig:pipeline}.
The understanding expert processes language instructions and visual content to predict subtasks and affordances, while the generation expert predicts future visual observations and robot actions.
The two branches are coupled through a standard Mixture-of-Transformers (MoT) architecture.
We initialize MachEmbodied-U0 from Lance~\cite{lance} and further train it to acquire robot-specific capabilities. 

\subsubsection{Understanding Expert}
We formulate subtask and affordance prediction as autoregressive text generation and train the understanding expert with a next-token cross-entropy objective. The resulting semantic visual tokens are combined with the task instruction in the shared multimodal sequence.

\noindent\textbf{Subtask Prediction.}
An overall task instruction can remain unchanged across several manipulation stages, while the immediate goal changes as the scene evolves. Following the use of semantic subtask supervision in $\pi_{0.5}$~\cite{pi05}, we ask the understanding expert to describe the local manipulation goal given the instruction and current observation. This encourages the expert to interpret the observed scene in the context of the overall task and distinguish the behaviors required at different stages.

\noindent\textbf{Affordance Prediction.}
For keyframes with affordance annotations, we model actionable spatial grounding using a similar question-answering formulation. The \texttt{Affordance Query} asks the model to identify a task-relevant target in the current image and predict its bounding box, followed by the intended interaction and the corresponding interaction point. When both annotations are available, the affordance follows the subtask in the same assistant response; otherwise, it is predicted alone. Samples without affordance annotations retain their original training objective.

\subsubsection{Generation Expert}
As shown in Fig.~\ref{fig:pipeline}, the generation expert jointly generates future visual dynamics and robot actions.
In the shared sequence, text tokens follow the current observation and robot-state tokens and precede the future visual and action tokens.
Causal attention prevents text tokens from attending to future targets.
During denoising, the generation expert uses shared multimodal attention to access representations of the input images and text from the understanding expert.
It generates RGB, depth, surface-normal, or optical-flow sequences together with continuous robot actions.

\textbf{Visual Dynamics Generation.}
The generation expert produces
visual sequences in the modality specified by the input prompt: RGB,
depth, surface normals, or optical flow. We represent geometry and
motion as three-channel visual targets~\cite{genception}, complementing
RGB appearance with scene structure, surface orientation, and inter-frame
motion. All modalities share a generation backbone and latent output
projection operating in the latent space of a common video VAE, whose
decoder reconstructs the predicted latents into the requested visual
modality. This formulation unifies RGB, geometry, and motion generation
without modality-specific prediction heads.

\textbf{Action Generation.}
For action generation, we represent each timestep in the predicted action sequence as one token.
At each denoising step, the generation expert processes the noisy action tokens together with the visual tokens through bidirectional self-attention.
An action-specific output head maps the resulting action representations to velocity predictions, which are used to update the noisy action sequence using explicit Euler steps.
This process proceeds in synchrony with visual generation to produce the final action sequence.

\subsection{Multi-rate Rotary Position Encoding}
We adopt the multimodal rotary position embedding (mRoPE) scheme~\cite{bai2025qwen25vltechnicalreport}, which encodes text positions and visual spatiotemporal coordinates through rotations of query and key representations.
Building on this scheme, we introduce Multi-rate Rotary Position Encoding (MRPE) to accommodate temporally subsampled video while retaining actions at their original sampling rate.
Each video latent frame corresponds to a chunk of consecutive actions, whose internal order would remain unspecified if they shared identical position indices.
We therefore retain the shared temporal position of the corresponding video latent frame, but assign each action a distinct height-axis index according to its order within the chunk, while keeping the width-axis index fixed.
This preserves temporal alignment between actions and video while explicitly encoding the position of each action within its chunk.

\subsection{Unified Multimodal Prompting}

Inspired by the use of contextual information in the prompts of $\pi_{0.7}$~\cite{intelligence2026pi07steerablegeneralistrobotic}, we design a unified multimodal prompt that combines the current visual observation and task instruction with control metadata and explicit prediction requirements. Control-mode and action-semantics fields specify the robot's control conventions, including relative or absolute commands and gripper encoding. Subtask and affordance queries define the requested understanding outputs, while a video-modality field specifies whether to generate RGB, depth, surface normals, or optical flow. Together, these fields provide the task and control context for joint understanding and generation. An example is shown below, with the shared system prompt omitted and visual embeddings represented by an observation placeholder:

\begin{tcolorbox}[
    colback=black!2,
    colframe=black!65,
    boxrule=0.9pt,
    arc=2mm,
    boxsep=0pt,
    left=10pt,
    right=10pt,
    top=8pt,
    bottom=8pt,
    fontupper=\ttfamily\small,
    before upper={\raggedright},
    before skip=8pt,
    after skip=8pt
]
\textless{}Current RGB observation\textgreater{} 
Task:pick up the cup.
Video Modality:rgb.
Control Mode:joint.
Action Semantics:joint delta (chunk-start); gripper absolute (0=open, 1=closed; pre-norm).
Subtask Query:Describe the current subtask based on the task and the current image.
Affordance Query:Identify a target relevant to carrying out the task in the current image. Give its name and bounding box. Output:\textless{}subtask\textgreater{}\textless{}affordance\textgreater{}
\end{tcolorbox}

\subsection{Unified Action Representation}

Embodiment diversity provides rich manipulation experience, but inconsistent control conventions can obscure the common structure across demonstrations. Similar numerical values may describe different physical quantities or even opposite control directions. 
We therefore adopt a unified action space for robot states and actions across robot embodiments, shown in Tab.~\ref{tab:unified-action-representation}. 

We implement dataset-specific adapters to align the physical meaning
of robot states and actions before statistical normalization.
These adapters reconcile units, coordinate conventions, and control
directions while preserving meaningful embodiment-specific differences.
For example, gripper states and actions are expressed in a shared
closing coordinate, where $0$ denotes fully open and $1$ fully closed.
This reverses and rescales Galaxea's native $[0,100]$ commands while
preserving AgiBot's action convention. AgiBot's millimetre-valued gripper states are calibrated separately
to the same closing coordinate.

Some sources additionally require kinematic conversion to make states
and actions physically consistent. Galaxea, for instance, records
mobile-base states as wheel-steering angles and wheel velocities,
but specifies actions as Cartesian velocities. We convert these
wheel-level states into body-frame planar velocities, giving states
and actions a common physical interpretation. Controls with distinct
physical meanings, such as torso position and velocity, remain explicitly distinguished.

\begin{table*}[t]
    \centering
    \small
    \setlength{\tabcolsep}{7pt}
    \renewcommand{\arraystretch}{1.12}
    \caption{
        Unified state and action representation. Dimensions are specified per arm or
        gripper.
    }
    \label{tab:unified-action-representation}
    \begin{tabular}{@{}lcl@{}}
        \toprule
        Component & Dim. & Shared representation \\
        \midrule
        Arm: joint control & 6 / 7
        & Joint configuration in radians \\
        Arm: EEF control & 9
        & Cartesian position in metres and 6D orientation \\
        Gripper & 1
        & Normalized closing coordinate: 0 = open, 1 = closed \\
        Chassis & 3
        & Body-frame planar velocity $(v_x,v_y,\omega_z)$, in m/s and rad/s \\
        Torso: AgiBot & 2
        & Pitch and lift position, in radians and metres \\
        Torso: Galaxea & 3
        & Planar Cartesian velocity $(v_x,v_z,\omega_y)$, in m/s and rad/s \\
        Head & 2
        & Yaw and pitch joint positions in radians \\
        \bottomrule
    \end{tabular}
\end{table*}
\section{Pretraining}

\subsection{Training Data Sampling}
\label{sec:data_sampling_recipe}

\begin{table*}[t]
    \centering
    \caption{Source-specific video--action temporal alignment.
    $H_a$ denotes the number of action steps, and $H_v$ the number
    of sampled video frames.}
    \label{tab:temporal-alignment}
    \small
    \setlength{\tabcolsep}{7pt}
    \renewcommand{\arraystretch}{1.08}
    \begin{tabular}{lcccc}
        \toprule
        Dataset source
        & Action (Hz)
        & Video (Hz)
        & $H_a$ / $H_v$
        & Horizon (s) \\
        \midrule

        AgiBot World Beta
        & 30 & 10 & 36 / 13 & 1.20 \\

        Galaxea Open World
        & 15 & 5 & 24 / 9 & 1.60 \\

        RoboMIND 2.0: AgileX / Mobile / ARK
        & 30 & 10 & 36 / 13 & 1.20 \\

        RoboMIND 2.0: Franka
        & 15 & 5 & 24 / 9 & 1.60 \\

        RoboCOIN: 30 Hz recordings
        & 30 & 10 & 36 / 13 & 1.20 \\

        RoboCOIN: 50 Hz recordings
        & 25 & 8.33 & 36 / 13 & 1.44 \\

        \bottomrule
    \end{tabular}
\end{table*}

Broad coverage of tasks and scene contexts provides a foundation for learning manipulation skills that generalizes beyond the pretraining data \cite{shi2026diversityneedscalablerobotic, pi05}. However, diversity in the collected corpus does not necessarily translate into diversity in training exposure. Under a fixed training budget, sampling in proportion to recorded timesteps can concentrate supervision on abundant tasks and long demonstrations, limiting exposure to less common behaviors. An effective sampling strategy should therefore preserve the breadth of the task repertoire while providing meaningful supervision across tasks.

We adopt global task-aware sampling to achieve this, allocating the training budget across source-specific task buckets rather than prescribing dataset-level proportions. Each eligible task receives an initial coverage allocation drawn from multiple episodes, ensuring that less abundant tasks also contribute to pretraining. With this coverage established, the remaining budget is then distributed through temperature-based sampling with weights
\[
w_k = \max(R_k, 1)^{\alpha},
\]
where $R_k$ denotes the task's remaining capacity in training blocks and $\alpha$ controls the degree of distribution smoothing. We set $\alpha=0.5$ to balance the contribution of less abundant tasks against the greater demonstration diversity available in larger task buckets. 
This sublinear weighting reduces the dominance of large task buckets while retaining a preference for tasks with more available data. 
Task allocations remain bounded by the available data capacity, avoiding aggressive oversampling of tasks with few demonstrations.
For tasks with insufficient capacity, we draw additional samples from egocentric datasets to supplement their coverage.
This combines broad task coverage with more balanced supervision without enforcing equal sample counts across tasks.
Dataset proportions emerge naturally from the resulting task-level allocation.
Using this strategy, we draw 120 million training samples from the eligible tasks in the curated data mixture.

\subsection{Video--Action Temporal Alignment}

Heterogeneous recording rates introduce a temporal mismatch when
combining robotic datasets: sequences with the same number of frames
may capture substantially different amounts of motion and span different
physical durations. We address this by jointly selecting source-specific
sampling strides and sequence lengths, keeping prediction windows within
approximately 1--2 seconds while sampling actions three times as densely
as video. This provides finer-grained control supervision without requiring
equally dense visual sequences.

Specifically, each interval between consecutive sampled video frames
corresponds to three action steps. For $H_a$ action steps and $H_v$ video
frames, including the anchor, we have
\[
H_a = 3(H_v - 1).
\]
Combined with the video VAE temporal compression, this
establishes a consistent correspondence of 12 action steps per predicted
video-latent step. The resulting representation preserves a common
cross-modal temporal structure across datasets while accommodating their
different recording rates. Tab.~\ref{tab:temporal-alignment} summarizes
the source-specific sampling settings.

\subsection{Understanding Expert Training}

We introduce two complementary language-supervision tasks to enhance task-grounded understanding. Subtask prediction connects the overall goal to the ongoing manipulation step, while affordance prediction identifies and localizes a task-relevant object or region in the current image. Together, they encourage the expert to associate semantic intent with concrete visual referents, providing richer supervision than continuous action targets alone. Both responses are trained with autoregressive cross-entropy \begin{equation}
\mathcal{L}_{\mathrm{und}}
=
-\frac{1}{T}\sum_{t=1}^{T}
\log p_{\theta}\!\left(y_t \mid y_{<t}, \mathbf{x}\right),
\end{equation}
where $\mathbf{x}$ comprises the current image, task instruction, and contextual metadata, and $\mathbf{y}=(y_1,\ldots,y_T)$ is the target response containing the available subtask and affordance labels. The loss is evaluated only on supervised response tokens, jointly learning task interpretation and visual grounding within a shared language-generation framework.

\subsection{Generation Expert Training}

The generation expert, initialized from the pretrained Lance~\cite{lance}
model, jointly denoises future visual latents and robot actions using
conditional Flow Matching~\cite{flow_matching}. Given the conditioning
context $\mathbf{c}$, we corrupt the clean visual target $\mathbf{z}$
and action trajectory $\mathbf{a}$ with independent standard Gaussian
noise at a shared flow time $\tau$. We sample
$\tau = \operatorname{sigmoid}(u+\log 4)$, where
$u \sim \mathcal{N}(0,1)$, yielding a shifted logit-normal distribution
that emphasizes higher noise levels:
\begin{equation}
\mathbf{z}_{\tau}
= (1-\tau)\mathbf{z}+\tau\boldsymbol{\epsilon}_{z},
\qquad
\mathbf{a}_{\tau}
= (1-\tau)\mathbf{a}+\tau\boldsymbol{\epsilon}_{a}.
\end{equation}
The noisy visual and action tokens are processed together by the
generation expert, allowing information exchange between modalities.
Separate output projections predict their respective velocity fields:
\begin{equation}
    \left(
    \widehat{\mathbf{u}}_{z},
    \widehat{\mathbf{u}}_{a}
    \right)
    =
    v_{\theta}
    \left(
    \mathbf{z}_{\tau},
    \mathbf{a}_{\tau},
    \tau;\mathbf{c}
    \right).
\end{equation}
The joint flow-matching objective is
\begin{equation}
    \mathcal{L}_{\mathrm{gen}}
    =
    \mathbb{E}
    \left[
    \lambda_z
    \left\|
    \widehat{\mathbf{u}}_{z}
    -(\boldsymbol{\epsilon}_{z}-\mathbf{z})
    \right\|_{\mathbf{M}_{z}}^2
    +
    \lambda_a
    \left\|
    \widehat{\mathbf{u}}_{a}
    -(\boldsymbol{\epsilon}_{a}-\mathbf{a})
    \right\|_{\mathbf{M}_{a}}^2
    \right],
\end{equation}
where $\|\cdot\|_{\mathbf{M}}^2$ denotes the mean squared error over valid
elements selected by mask $\mathbf{M}$.

We adopt joint denoising to preserve the unified understanding and
generation architecture of our model. Language and visual understanding
provide the conditioning context, while the same generation expert
jointly processes noisy action tokens and future visual latents.
Introducing a separate action denoiser would decouple action generation
from this shared generative process and weaken the architectural
unification of understanding, visual prediction, and robot control.
Instead, joint denoising incorporates actions directly into the
multimodal generation process, allowing action and visual predictions
to interact throughout denoising. Modality-specific output projections
accommodate their different representations while preserving a shared
generation backbone.

For visual dynamics generation, selected future RGB targets are replaced with
depth, surface normals, or optical flow, with the target
modality specified by the language instruction. All modalities share
the same video VAE and generation expert, while the current RGB
observation and action targets remain unchanged. This allows geometric
prediction and action generation to be jointly optimized under the
same flow-matching objective.

Beyond joint video and action generation, the architecture supports training
on forward and inverse dynamics tasks~\cite{cosmos3} by varying the conditioning
inputs and prediction targets. Forward dynamics predicts future visual latents
conditioned on the current observation and an action trajectory, whereas
inverse dynamics predicts actions conditioned on the current and future
visual latents:

\begin{align}
    \text{Forward dynamics:}\quad
    &(\mathbf{c},\mathbf{a})\longrightarrow\mathbf{z},
    \\
    \text{Inverse dynamics:}\quad
    &(\mathbf{c},\mathbf{z})\longrightarrow\mathbf{a},
    \\
    \text{Joint policy prediction:}\quad
    &\mathbf{c}\longrightarrow(\mathbf{z},\mathbf{a}).
\end{align}

The overall training objective combines the generation expert's joint
flow-matching loss with the understanding expert's autoregressive
cross-entropy loss:
\begin{equation}
    \mathcal{L}_{\mathrm{total}}
    = \mathcal{L}_{\mathrm{gen}}
    + \lambda_{\mathrm{und}}\mathcal{L}_{\mathrm{und}},
    \label{eq:total-training-loss}
\end{equation}
where $\lambda_{\mathrm{und}}$ controls the relative contribution of
understanding supervision, while $\lambda_z$ and $\lambda_a$ in
$\mathcal{L}_{\mathrm{gen}}$ balance visual and action generation.
This objective jointly optimizes task-grounded understanding, future
visual prediction, and robot action generation.

\section{Training Infrastructure}
\label{sec:training-infra}

To train efficiently on our large-scale pretraining data mixture, we design our training infrastructure to address two costly recurring operations: dataset initialization and video data loading. 
The data-initialization optimizations reduce the initialization time of the full pretraining mixture from approximately one hour to less than ten minutes, and the video data loading strategy reduces overall training time by more than $30\%$ in our training platform.

\subsection{Efficient Dataset Initialization}
\label{sec:infra-manifest-initialization}

\noindent\textbf{Precomputed episode manifests.}
Initializing a large dataset mixture can be slow because it requires traversing source directories and inspecting episode files, particularly when metadata must be retrieved over the network. We move this discovery process to the data-preparation stage and record episode identifiers, lengths, task associations, and source-specific filtering information in persistent manifests. At training time, the episode catalog is reconstructed directly from these manifests, eliminating repeated file-system inspection.

\noindent\textbf{Reusable Subtask Index.}
We precompute a reusable metadata index containing each subtask’s episode identifier, temporal boundaries, and language label. Training runs load this index directly, avoiding repeated parsing and alignment of the original annotations during initialization. Rather than storing experiment-specific training windows, the index preserves the complete annotated intervals, from which eligible anchors are derived according to the configured video and action offsets. It can therefore be reused across experiments with different prediction horizons.

\noindent\textbf{Lightweight Data Readers.}
Constructing a full Dataset object for every constituent dataset introduces substantial initialization overhead, particularly for large sources distributed across hundreds of LeRobot directories. To reduce this overhead, we implement lightweight LeRobot readers that use precomputed metadata to access the underlying files without repeating the full initialization procedure.
We additionally support direct access to native data formats through source-specific readers, avoiding the cost of converting large datasets into a common storage format.

\begin{figure*}[t]
    \centering
    \includegraphics[width=\textwidth]{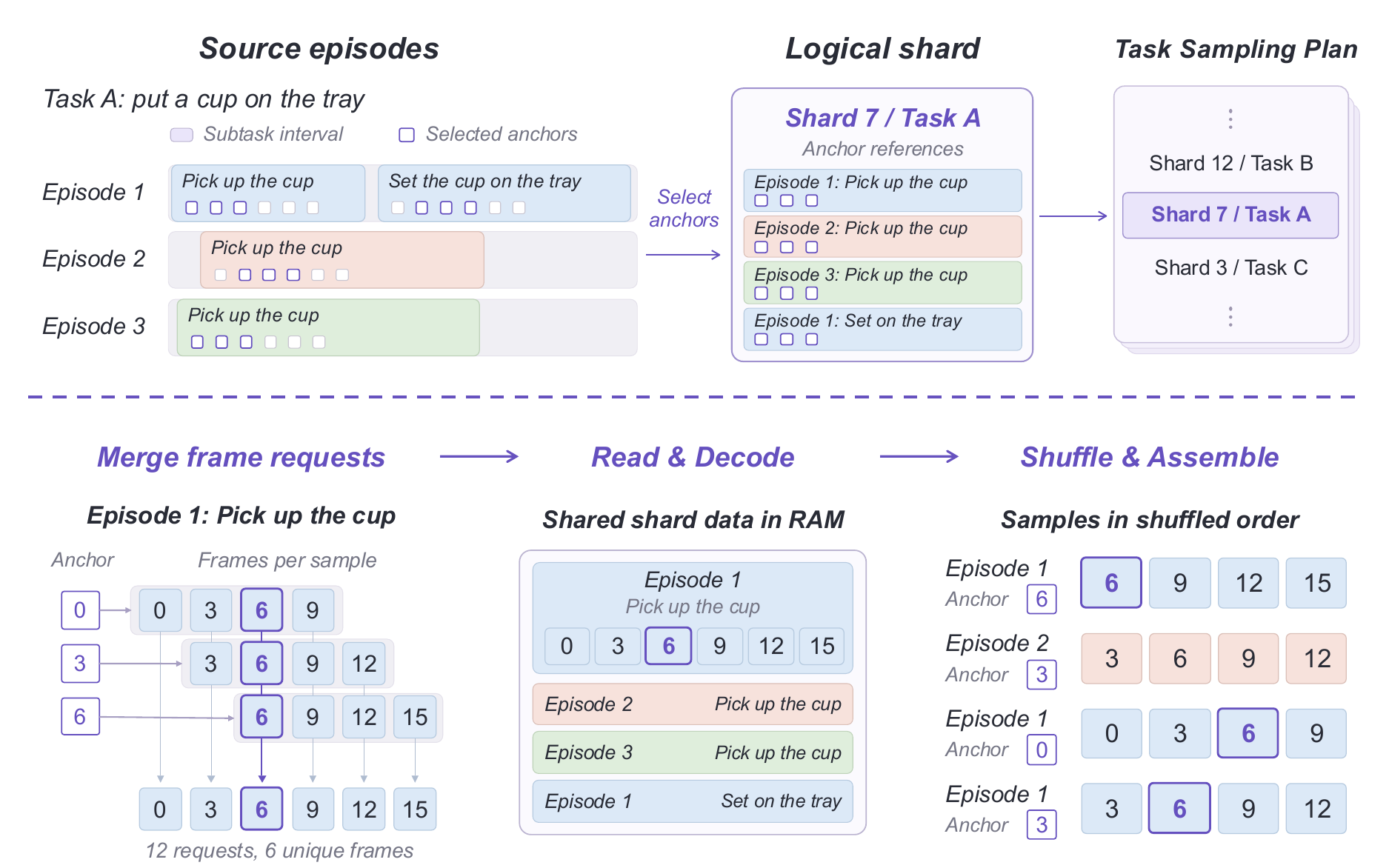}
    \caption{
    \textbf{Shard-based sampling and loading.}
    Top: selected anchors from subtask intervals across episodes of the same
    task are referenced by a logical shard from the global
    task sampling plan.
    Bottom: overlapping frame requests are merged, decoded into shared
    memory buffers, and assembled into samples in shuffled anchor order.
    }
    \label{fig:shard-sampling-loading}
\end{figure*}

\subsection{Logical Shards}
\label{sec:infra-window-sharding}

Video-based model training incurs substantial I/O and video decoding overhead for the data pipeline.
We organize training samples into logical shards, which serve as the basic units of allocation for the global task sampler and of grouped loading for data workers.
This preserves temporal locality for efficient video decoding while allowing the sampler to control coverage across tasks and episodes. 

The upper part of Fig.~\ref{fig:shard-sampling-loading} shows how demonstrations are organized into logical shards. Each training sample starts at an \emph{anchor}, whose complete video/action window must remain within the same subtask interval and episode.
To construct a shard, we cycle through episodes of one task, taking nearby anchors from each subtask interval until the shard contains the configured number of anchors. These shards form the units of the global task sampling plan. Following the task-level allocation described above, the plan selects
shards without replacement and shuffles them into a reproducible schedule for distributed workers.

The lower part of Fig.~\ref{fig:shard-sampling-loading} shows how a worker loads a scheduled shard. Rather than loading each sample independently, a worker merges overlapping frame requests from nearby anchors within the same subtask interval. In the example, twelve requests reduce to six unique frames, which are decoded and kept in memory alongside the state/action data. The worker then shuffles the anchors and assembles their samples from the shared data.
Training samples are thus drawn from different episode segments in shuffled order, while overlapping samples continue to share the same decoded frames.

\subsection{Video Decoding and CPU Offloading}
\label{sec:infra-decoding}

We cap the number of nearby anchors processed per decoding job, restricting its video span to those anchors and their required future frames rather than the entire episode.
This balances frame reuse against per-job memory demand. A separate concurrency limit controls aggregate CPU and memory usage, while asynchronous preparation of the next shard overlaps
decoding with training.

To accommodate decoding workloads that exceed the CPU capacity of GPU nodes, we also support offloading data preparation to separate CPU jobs.
These jobs read and decode data ahead of training and place the prepared payloads in a shared cache.
Training workers retrieve cached payloads without performing local video decoding on cache hits.
The CPU jobs follow the training sample schedule and maintain a bounded lead over consumption, preventing excessive accumulation of prepared data.
This separation allows decoding capacity to scale independently of GPU resources and reduces training time significantly.

\begin{table*}[t]
    \centering
    \caption{Results on RoboDojo-Sim.
    SR and Score are reported on a 0--100 scale, with higher values better. Bold indicates the best result within each model group.}
    \label{tab:robodojo_sim_leaderboard}
    \small
    \setlength{\tabcolsep}{3pt}
    \renewcommand{\arraystretch}{1.1}
    \resizebox{\textwidth}{!}{%
    \begin{tabular}{l*{14}{c}}
        \toprule
        \multirow{2}{*}{Model}
        & \multicolumn{2}{c}{Gen-Std}
        & \multicolumn{2}{c}{Gen-Rand}
        & \multicolumn{2}{c}{Precision}
        & \multicolumn{2}{c}{Long-Horizon}
        & \multicolumn{2}{c}{Memory}
        & \multicolumn{2}{c}{Open}
        & \multicolumn{2}{c}{Overall} \\
        \cmidrule(lr){2-3}\cmidrule(lr){4-5}\cmidrule(lr){6-7}
        \cmidrule(lr){8-9}\cmidrule(lr){10-11}\cmidrule(lr){12-13}
        \cmidrule(lr){14-15}
        & SR & Score & SR & Score & SR & Score
        & SR & Score & SR & Score & SR & Score & SR & Score \\

        \midrule
        \rowcolor[gray]{0.93}
        \multicolumn{15}{c}{\textbf{VLA}} \\
        \midrule

        StarVLA~\cite{starvla}
        & 4.67 & 7.54 & 0.00 & 0.33 & 4.33 & 9.90
        & 6.50 & 14.15 & 2.44 & 3.34
        & 0.58 & 0.67 & 3.24 & 6.40 \\

        X-VLA~\cite{xvla}
        & 12.22 & 17.90 & 1.33 & 3.04 & 12.00 & 18.32
        & 9.75 & 16.53 & 3.56 & 4.76
        & 0.50 & 0.55 & 6.52 & 10.13 \\

        $\pi_{0.5}$~\cite{pi05}
        & 14.89 & 20.93 & 1.44 & 5.82 & 5.50 & 12.40
        & 14.67 & 23.54 & 4.67 & 5.89
        & 1.67 & 1.98 & 6.93 & 11.44 \\

        Spatial Forcing~\cite{spatialforcing}
        & 14.89 & 21.25 & 3.78 & 6.98 & 10.58 & 17.32
        & 14.58 & 23.26 & 4.11 & 5.43
        & 1.58 & 1.78 & 8.04 & 12.38 \\

        Hy-Embodied-0.5-VLA~\cite{hyembodied05}
        & 16.56 & 21.98 & 0.22 & 1.57 & 8.00 & 13.81
        & 14.92 & 25.74 & 12.11 & 13.37
        & 0.58 & 0.65 & 8.80 & 13.07 \\

        Xiaomi-Robotics-1~\cite{xiaomi1}
        & \textbf{28.00} & \textbf{35.65}
        & \textbf{6.00} & \textbf{11.44}
        & 18.83 & 26.69 & 23.67 & 38.39 & 6.56 & 7.81
        & \textbf{3.58} & \textbf{3.94} & 13.93 & 20.07 \\

        GalaxeaVLA (G0.5)~\cite{galaxea05}
        & 21.44 & 27.72 & 4.22 & 9.20
        & \textbf{20.42} & \textbf{28.25}
        & \textbf{32.25} & \textbf{44.12}
        & 7.33 & 8.61 & 1.58 & 1.73 & 14.88 & 20.23 \\

        DM0.5~\cite{dm05}
        & 17.89 & 23.49 & 4.00 & 8.06 & 16.75 & 24.82
        & 19.50 & 33.70 & \textbf{47.44} & \textbf{47.74}
        & 2.08 & 2.43 & \textbf{19.34} & \textbf{24.90} \\

        \midrule
        \rowcolor[gray]{0.93}
        \multicolumn{15}{c}{\textbf{WAM}} \\
        \midrule

        Fast-WAM~\cite{fastwam}
        & 2.11 & 4.33 & 0.11 & 0.34 & 0.00 & 1.96
        & 5.17 & 9.14 & 3.44 & 3.55
        & 0.42 & 0.42 & 2.03 & 3.48 \\

        AHA-WAM~\cite{ahawam}
        & 6.22 & 10.32 & 0.33 & 1.26 & 2.42 & 5.86
        & 2.67 & 8.61 & 2.78 & 2.97
        & 0.83 & 0.88 & 2.39 & 4.82 \\

        GigaWorld-Policy-0~\cite{gigaworldpolicy}
        & 5.78 & 10.28 & 0.00 & 0.41 & 1.83 & 6.15
        & 8.92 & 15.51 & 2.22 & 3.46
        & 0.50 & 0.54 & 3.27 & 6.20 \\

        X-WAM~\cite{xwam}
        & 5.33 & 11.24 & 1.33 & 3.54 & 1.83 & 6.72
        & 9.08 & 17.47 & 4.67 & 6.32
        & 0.25 & 0.57 & 3.83 & 7.69 \\

        OpenWAM-$\alpha$~\cite{openwam}
        & \textbf{25.56} & \textbf{33.16}
        & \textbf{4.11} & \textbf{8.26}
        & 9.25 & 18.45 & \textbf{25.33} & 34.93
        & \textbf{9.11} & \textbf{10.41}
        & \textbf{1.08} & \textbf{1.41}
        & \textbf{11.92} & 17.18 \\

        \midrule
        \textbf{ME-U0 (ours)}
        & 19.11 & 28.10 & 1.33 & 6.96
        & \textbf{15.42} & \textbf{23.95}
        & 22.33 & \textbf{36.98}
        & 7.00 & 8.42 & 0.92 & \textbf{1.41}
        & 11.18 & \textbf{17.66} \\
        \bottomrule
    \end{tabular}%
    }
\end{table*}

\section{Experimental Results}
\label{sec:experiments}

\subsection{Post-Training Recipe}
We separately post-train ME-U0 on RoboDojo~\cite{robodojo} and LIBERO~\cite{libero}, initializing both runs from the same pretrained checkpoint.
Both runs use 368 PPU810E accelerators with a per-device batch size of 2. 
We use an action horizon of 24 with end-effector pose control for LIBERO,
and an action horizon of 48 with delta joint-position control for
RoboDojo.


It is worth noting that neither RoboDojo nor LIBERO provides subtask or affordance annotations, or auxiliary supervision for video geometry and motion.
For fair comparisons, we do not augment either dataset with these additional supervision signals during post-training.
Despite the potential degradation of pretrained representations under this reduced-supervision post-training setting, ME-U0 achieves strong performance on both benchmarks, highlighting the robustness of its pretrained representations and their effectiveness for downstream adaptation.

\begin{table*}[!t]
    \centering
    \small
    \renewcommand{\arraystretch}{1.0}
    \setlength{\abovecaptionskip}{4pt}
    \setlength{\belowcaptionskip}{4pt}

    \caption{Post-training results on standard LIBERO.}
    \label{tab:libero-results}

    \setlength{\tabcolsep}{6.5pt}
    \begin{tabular*}{\textwidth}{
        @{\hspace{\tabcolsep}\extracolsep{\fill}}
        l*{5}{c}
        @{\hspace{\tabcolsep}}
    }
        \toprule
        Model & Spatial & Object & Goal & Long & Avg. \\

        \midrule
        \rowcolor[gray]{0.93}
        \multicolumn{6}{c}{\textbf{VLA}} \\
        \midrule

        $\pi_{0.5}$~\cite{pi05}
        & \textbf{98.8} & 98.2 & 98.0 & 92.4 & 96.9 \\

        OpenVLA-OFT~\cite{openvlaoft}
        & 97.6 & 98.4 & 97.9 & 94.5 & 97.1 \\

        StarVLA~\cite{starvla}
        & 97.8 & 98.6 & 96.2 & 93.8 & 96.6 \\

        ABot-M0~\cite{abotm0}
        & \textbf{98.8} & \textbf{99.8} & \textbf{99.0}
        & 96.6 & \textbf{98.6} \\

        X-VLA~\cite{xvla}
        & 98.2 & 98.6 & 97.8 & \textbf{97.6} & 98.1 \\

        \midrule
        \rowcolor[gray]{0.93}
        \multicolumn{6}{c}{\textbf{WAM}} \\
        \midrule

        Fast-WAM~\cite{fastwam}
        & 98.2 & \textbf{100.0} & 97.0 & 95.2 & 97.6 \\

        Motus~\cite{motus}
        & 96.8 & 99.8 & 96.6 & 97.6 & 97.7 \\

        ImageWAM~\cite{imagewam}
        & 97.2 & 99.2 & 98.8 & 98.4 & 98.4 \\

        LingBot-VA~\cite{li2026causal}
        & 98.5 & 99.6 & 97.2 & 98.5 & 98.5 \\

        OpenWAM-$\alpha$~\cite{openwam}
        & \textbf{99.6} & 99.6 & \textbf{99.8}
        & 98.2 & \textbf{99.3} \\

        \midrule
        \textbf{ME-U0 (ours)}
        & 98.4 & 99.8 & 99.0 & \textbf{98.6} & 99.0 \\
        \bottomrule
    \end{tabular*}

    \par\vspace{8pt}

    \caption{Robustness on LIBERO-Plus. Success rates (\%) are reported
    across seven perturbation categories~\cite{libero_plus}.}
    \label{tab:libero-plus-results}

    \setlength{\tabcolsep}{3.7pt}
    \begin{tabular*}{\textwidth}{
        @{\hspace{\tabcolsep}\extracolsep{\fill}}
        l*{8}{c}
        @{\hspace{\tabcolsep}}
    }
        \toprule
        Model & Camera & Robot & Language & Light
        & Background & Noise & Layout & Avg. \\

        \midrule
        \rowcolor[gray]{0.93}
        \multicolumn{9}{c}{\textbf{VLA}} \\
        \midrule

        StarVLA~\cite{starvla}
        & 52.5 & 49.8 & \textbf{88.5} & 95.7
        & 95.7 & 73.0 & 76.9 & 74.1 \\

        ABot-M0~\cite{abotm0}
        & 60.4 & 67.9 & 86.4 & 96.2
        & 91.6 & 86.4 & 82.6 & 80.5 \\

        $\pi_{0.5}$~\cite{pi05}
        & 78.4 & 73.6 & 80.8 & 96.2
        & 94.1 & 89.0 & 84.5 & 84.4 \\

        Qwen-RobotManip~\cite{yuan2026qwen}
        & \textbf{87.2} & \textbf{75.5} & 85.6 & \textbf{96.6}
        & \textbf{97.7} & \textbf{97.7} & \textbf{87.3}
        & \textbf{89.0} \\

        \midrule
        \rowcolor[gray]{0.93}
        \multicolumn{9}{c}{\textbf{WAM}} \\
        \midrule

        Fast-WAM~\cite{fastwam}
        & 16.4 & 44.5 & 68.9 & 78.2
        & 53.7 & 37.7 & 60.7 & 51.5 \\

        Being-H0.7~\cite{beingh07}
        & \textbf{82.0} & 59.0 & 82.8 & 97.8
        & \textbf{90.0} & 93.5 & \textbf{88.5} & 82.1 \\

        Cosmos-Policy~\cite{cosmospolicy}
        & 75.8 & 63.3 & 81.7 & 96.5
        & 88.9 & 92.7 & 82.2 & 82.2 \\

        ImageWAM~\cite{imagewam}
        & 80.8 & 50.3 & \textbf{91.4} & \textbf{98.1}
        & 85.5 & \textbf{93.8} & 80.5 & \textbf{83.1} \\

        OpenWAM-$\alpha$~\cite{openwam}
        & 33.8 & \textbf{76.1} & 88.0 & 97.0
        & 87.1 & 39.8 & 77.5 & 69.2 \\

        \midrule
        \textbf{ME-U0 (ours)}
        & 70.2 & 75.7 & 90.4 & 97.6 & 81.9 & 81.3 & 84.5 & 82.5 \\
        \bottomrule
    \end{tabular*}
\end{table*}

\subsection{Simulation Benchmark Evaluation}
\subsubsection{RoboDojo}
We evaluate ME-U0 on the 42 tasks of RoboDojo~\cite{robodojo}
across generalization, precision, long-horizon execution, memory, and
open-vocabulary instruction following. For each of three evaluation seeds,
we run 25 standard and 25 randomized episodes for each of the 12
generalization tasks, and 50 episodes for each of the remaining 30 tasks. We report success rate (SR) and the
benchmark's process score, which gives partial credit for task progress.
Overall results
average the five capability dimensions equally, with Gen-Std and Gen-Rand
first combined into the generalization dimension.

Tab.~\ref{tab:robodojo_sim_leaderboard} compares ME-U0 with representative
VLA and WAM baselines. ME-U0 achieves the strongest overall performance
among the evaluated WAMs, reaching an overall process score of 17.66 and
an SR of 11.18\%. The advantage is particularly evident in precision and
long-horizon execution, where ME-U0 obtains process scores of 23.95 and
36.98, respectively. These results suggest that grounding joint
visual--action dynamics with subtask and affordance context supports both
fine-grained interaction and extended task execution. ME-U0 also remains
competitive with established VLA baselines despite being pretrained on
only approximately 4,200 hours of curated robotic and egocentric demonstrations, demonstrating competitive downstream transfer from a comparatively moderate pretraining corpus. Memory-dependent tasks
remain comparatively challenging, with a process score of 8.42 and an SR
of 7.00\%, likely because the current model does not explicitly retain
observation history. Incorporating temporal context and memory mechanisms
could further improve performance when task-relevant information is no
longer visible in the current observation.

\subsubsection{LIBERO/LIBERO-Plus}

\label{sec:libero-results}

We evaluate a single LIBERO-adapted ME-U0 checkpoint on the four standard
LIBERO suites and LIBERO-Plus, which extends the same task families along
seven controlled perturbation dimensions~\cite{libero_plus}. Standard
LIBERO evaluates 50 initial states for each of its 40 tasks, resulting in
2{,}000 episodes, whereas LIBERO-Plus evaluates one episode for each of
10{,}030 perturbed task instances. We use the same checkpoint for both
benchmarks without LIBERO-Plus-specific adaptation and aggregate success
over episodes. The overall LIBERO-Plus result is therefore weighted by
the number of task instances in each perturbation category. Published
baseline results are transcribed from the cited reports for reference
and were not rerun in our evaluation environment.

As Tab.~\ref{tab:libero-results} shows, ME-U0 achieves an average success rate of 99.0\% on LIBERO, including 98.4\% on Spatial, 99.8\% on Object, 99.0\% on Goal, and 98.6\% on Long, demonstrating consistently near-ceiling performance across all four suites. Without additional adaptation, the same checkpoint achieves an overall success rate of 82.5\% on LIBERO-Plus, as reported in Tab.~\ref{tab:libero-plus-results}. Performance remains particularly strong under lighting and language perturbations, with success rates of 97.6\% and 90.4\%, respectively. Camera-viewpoint changes and variations in robot initial states are the most challenging categories, yielding 70.2\% and 75.7\%. Success rates under background textures, sensor noise, and object layouts are 81.9\%, 81.3\%, and 84.5\%, respectively. These results show that LIBERO-Plus provides a more discriminative assessment than the nearly saturated standard benchmark and identify robustness to viewpoint changes and robot initial states as areas for further improvement.

\subsection{Real-World Deployment}
\begin{figure*}[t]
    \centering
    \includegraphics[width=\textwidth]{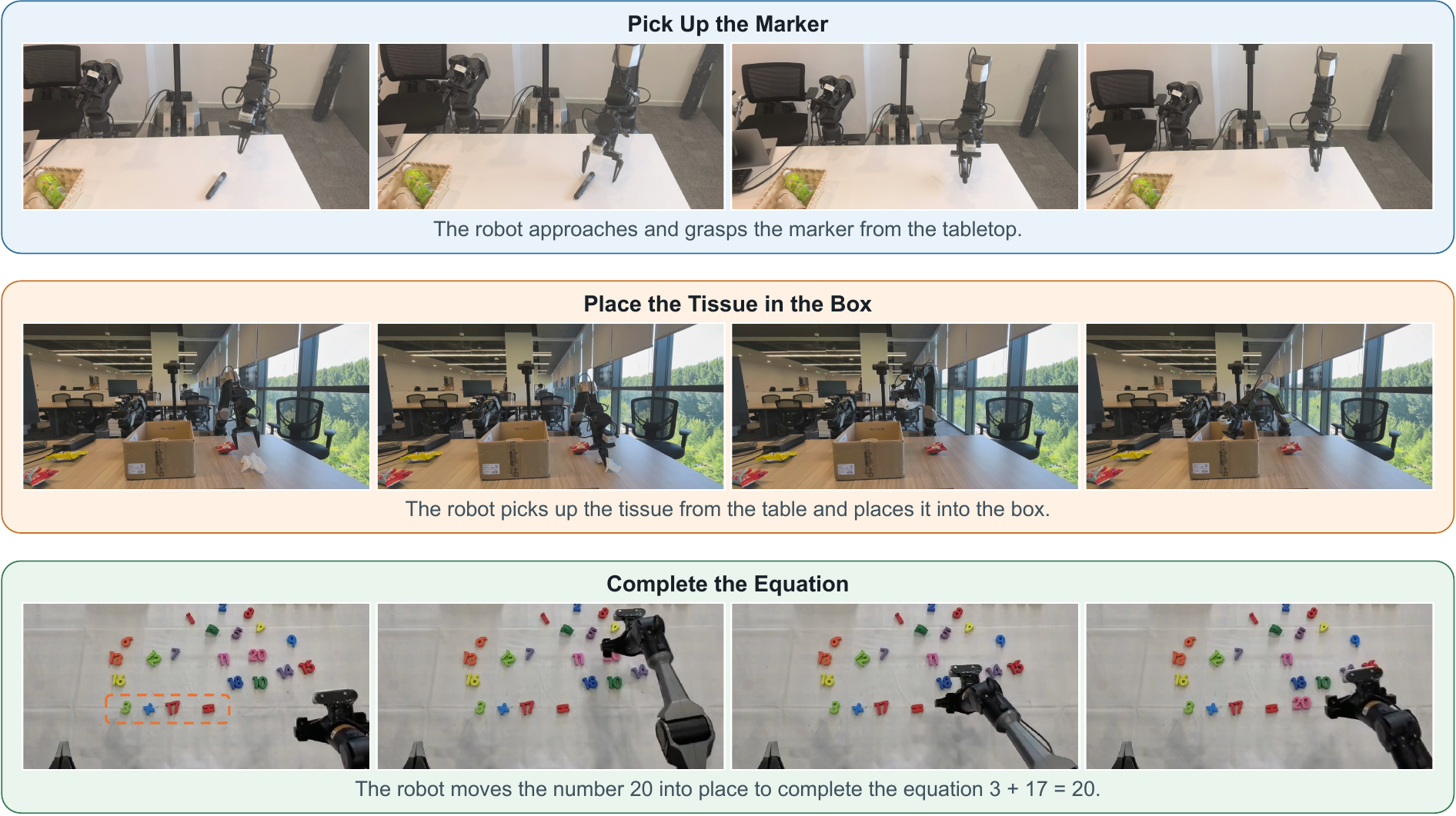}
    \caption{Real-world deployment demonstration of ME-U0.}
    \label{fig:real_world_grasping}
\end{figure*}

\begin{figure*}[!tp]
    \centering
    \includegraphics[width=\textwidth]{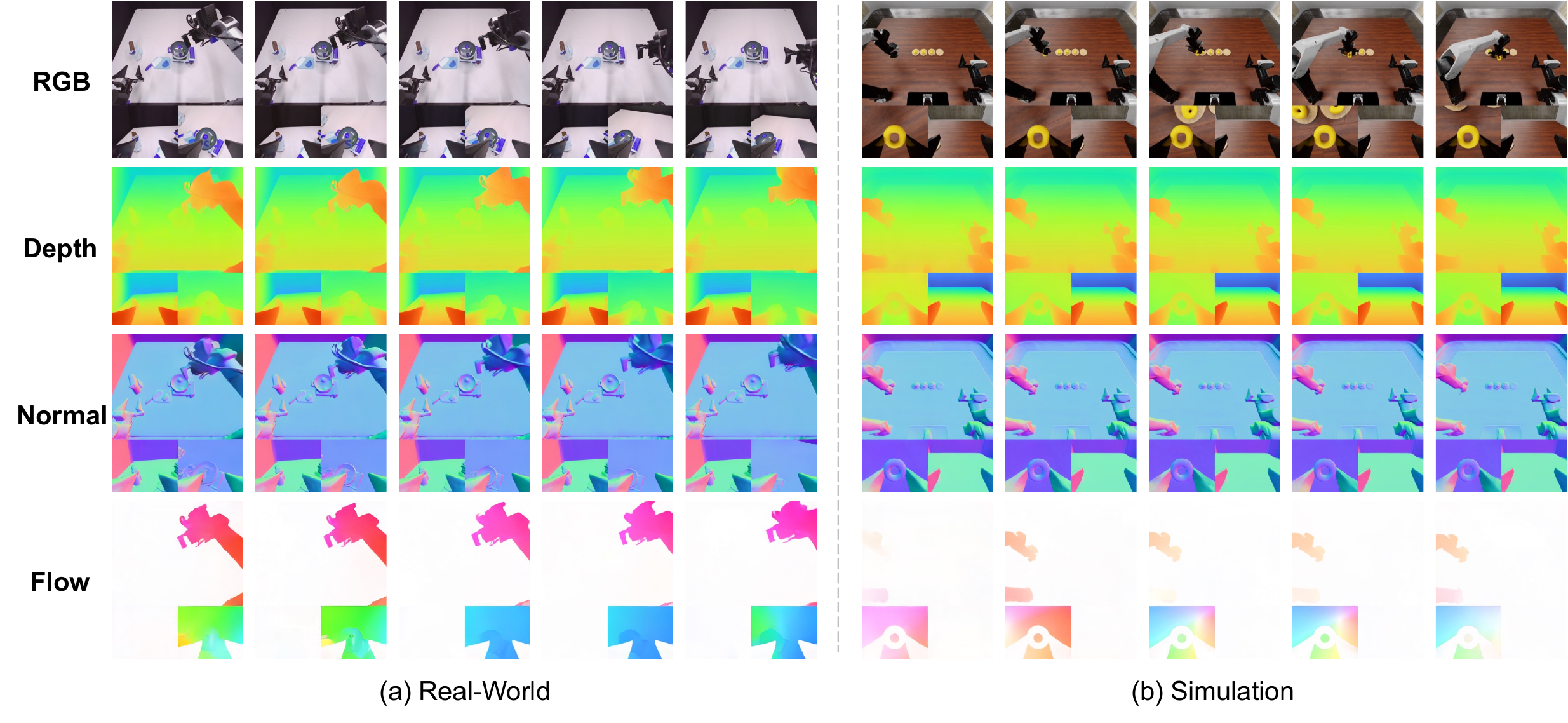}
    \caption{Zero-shot visual dynamics generation on RoboDojo
    (a) Real-World and (b) Simulation data. Each panel shows RGB
    observations alongside generated depth, surface normals, and
    optical flow at five time steps.}
    \label{fig:genception_sim_real}
\end{figure*}
\begin{figure*}[!t]
    \centering
    \includegraphics[width=\textwidth]{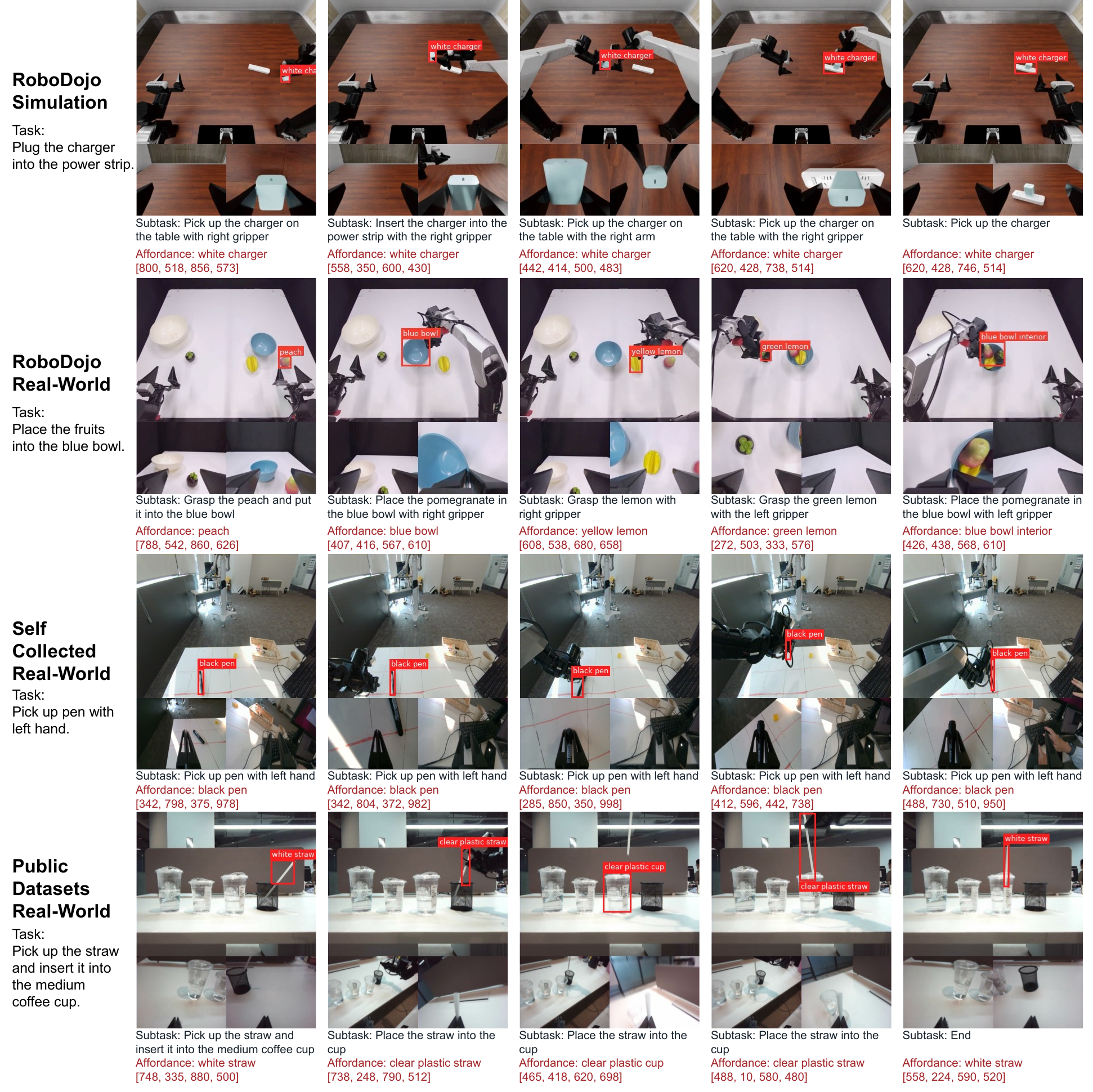}
    \caption{\textbf{Zero-shot affordance and subtask prediction.}
    Rows show RoboDojo simulation, RoboDojo real-world, self-collected, and RDT-1B~\cite{rdt} data.
    Task instructions appear on the left, with predicted subtasks and affordances below each image. Red boxes mark predicted task-relevant targets.}
    \label{fig:affordance}
\end{figure*}

We evaluate ME-U0 through direct deployment on real-world manipulation
tasks across multiple robot platforms, with model inference performed
on a remote server. The pretrained checkpoint is used without additional
post-training or deployment-specific adaptation. Since these platforms
are represented in the pretraining corpus, this evaluation examines
deployment on familiar embodiments rather than generalization to
unseen robot platforms.
Fig.~\ref{fig:real_world_grasping} presents three representative
instruction-guided tasks: picking up a marker, placing a tissue into
a box, and positioning the number 20 to complete the equation
$3 + 17 = 20$. Spanning grasping, pick-and-place manipulation, and
semantically grounded object placement, these demonstrations provide
qualitative evidence of ME-U0's ability to translate task instructions
into executable actions in real-world settings.

\subsection{Zero-Shot Analysis}
To assess the generalization capabilities of ME-U0, we conduct three zero-shot experiments using its pretrained weights directly, without any additional post-training: visual dynamics generation, affordance prediction, and subtask prediction.
All evaluation samples are drawn from data excluded from our pretraining corpus and are used solely for evaluation.
\subsubsection{Visual Dynamics}
We evaluate visual dynamics generation on samples drawn from the simulation and real-world training splits of RoboDojo~\cite{robodojo}. 
Neither split overlaps with our pretraining corpus, and the samples are used exclusively for evaluation without benchmark-specific fine-tuning.
Fig.~\ref{fig:genception_sim_real} shows qualitative results for both domains. 
ME-U0 generates depth maps, surface normals, and optical flow that capture meaningful scene structure and motion across real-world and simulated observations, demonstrating transfer to previously unseen data. 
Qualitatively, predictions on real-world samples are slightly more consistent than those on simulated samples. 
We hypothesize that this difference arises from the domain gap between simulated imagery and the real-world observations encountered during pretraining.
\subsubsection{Affordance}
We evaluate affordance prediction on samples from the simulation and real-world training splits of RoboDojo~\cite{robodojo}.
We also include real-world data: self-collected observations and samples from the dataset released with RDT-1B~\cite{rdt}.
Given the current observation and task instruction, the understanding expert identifies a task-relevant object or region and predicts its bounding box.
Fig.~\ref{fig:affordance} shows the predicted target names and bounding boxes in this setting.
These examples include objects such as a charger, pieces of fruit, a pen, and a straw, as well as regions inside a bowl and on a table.
These qualitative results suggest that ME-U0 can identify and localize task-relevant objects and regions in both simulated and real-world scenes.

\subsubsection{Subtask}
%
We evaluate subtask prediction on the same samples used for affordance prediction.
Given the overall task instruction and current observation, the understanding expert describes what the robot should do at the current stage of the task.
Fig.~\ref{fig:affordance} also demonstrates the predicted subtasks and their corresponding affordance targets. 
Each subtask specifies the action to perform at the current stage, while the corresponding affordance target identifies the task-relevant object or region.

\section{Conclusion}
\label{sec:conclusion}

We presented \textbf{MachEmbodied-U0 (ME-U0)}, a unified embodied
foundation model that connects task-grounded understanding,
geometry- and motion-aware visual dynamics, and continuous action
generation. Through a Mixture-of-Transformers architecture, subtask
prediction and affordance grounding provide the generation process with
semantic and spatial context, while future RGB, depth, surface normals,
optical flow, and continuous actions are jointly modeled through flow
matching. Pretraining on approximately 4,200 hours of curated robotic and
egocentric demonstrations, with the robotic corpus spanning six
embodiments, enables ME-U0 to achieve competitive
performance on RoboDojo, LIBERO, and LIBERO-Plus, with additional
validation on real-world manipulation tasks. The model also retains
zero-shot subtask prediction, affordance grounding, and visual-dynamics
generation without corresponding downstream supervision.

Future work will extend ME-U0 to broader distributions of tasks,
environments, and embodiments, while strengthening long-term memory
and in-context adaptation for more general embodied interaction.

\section{Contributions and Acknowledgments}
\label{sec:contributions}

\parab{Contributors.}
\noindent

\textbf{Data:}
Wenfu Wang$^{*}$,
Kunsong Shi$^{*}$,
Yiren Zhang,
Jingke Wang$^{\dagger}$,
Yueran Zhao$^{\dagger}$,
Xuancheng Zhang,
Nanfei Ye.

\noindent
\textbf{Base Model:}
Haoran Wen$^{*}$,
Wenfu Wang$^{*}$,
Kunsong Shi$^{*}$,
Wancheng Feng$^{*}$.

\noindent
\textbf{Training Infrastructure:}
Wenfu Wang$^{*}$,
Jingke Wang$^{*\dagger}$,
Xingru Chen,
Zhaohong Sun,
Chengmin Yang.

\noindent
\textbf{Evaluation:}
Jingke Wang$^{*\dagger}$,
Kunsong Shi$^{*}$,
Wancheng Feng,
Zikang Yu,
Penghao Bi,
Yueran Zhao,
Jia Shi.

\noindent
\textbf{Project Lead:}
Haoran Wen,
Yu Liu.

\noindent
\textbf{Advisors:}
Kun Zhan,
Yan Xie.

\vspace{6pt}
\parab{Acknowledgments.}
We thank Xuyao Huang$^{\dagger}$,
Runsheng Wang,
Jiahao Gu,
Mingcui Wang,
Yue Ma,
Danlu Dong,
Xueyang Zhang,
Mofan Zhou,
Yuying Chen
for their valuable support and contributions.

\vspace{6pt}
\noindent
{\small
$^{*}$Core contributors.\quad
$^{\dagger}$Interns.\quad
}

\bibliography{references}

@inproceedings{xvla,
  title={X-vla: Soft-prompted transformer as scalable cross-embodiment vision-language-action model},
  author={Zheng, Jinliang and Li, Jianxiong and Wang, Zhihao and Liu, Dongxiu and Kang, Xirui and Feng, Yuchun and Zheng, Yinan and Zou, Jiayin and Chen, Yilun and Zeng, Jia and others},
  booktitle={International Conference on Learning Representations},
  volume={2026},
  pages={60580--60606},
  year={2026}
}

@article{ahawam,
  title={AHA-WAM: Asynchronous Horizon-Adaptive World-Action Modeling with Observation-Guided Context Routing},
  author={Cai, Jisong and Ling, Long and Chu, Shiwei and Liu, Zhongshan and Kang, Jiayue and Liang, Zhixuan and Xu, Wenjie and Mao, Yinan and Zhang, Weinan and Yang, Xiaokang and others},
  journal={arXiv preprint arXiv:2606.09811},
  year={2026}
}

@article{openwam,
  title={OpenWAM: An Open, Modular Exploration Towards Systematic World-Action Model Pretraining},
  author={Wang, Yuran and Huang, Siqiao and Li, Mingleyang and Zhang, Chenhao and Liang, Jiaqi and Jin, Weiyang and Chen, Yue and Chi, Xuemin and Zhou, Donghao and Yu, Qize and others},
  journal={arXiv preprint arXiv:2609.07398},
  year={2026}
}

@article{hyembodied05,
  title={Hy-embodied-0.5-vla: From vision-language-action models to a real-world robot learning stack},
  author={Zhang, He and Xiang, Lingzhu and Lin, Haitao and Huang, Zeyu and Wang, Minghui and Zhong, Dingyan and Dong, Yubo and Wu, Yihao and Rao, Yongming and Zhang, Dongsheng and others},
  journal={arXiv preprint arXiv:2606.14409},
  year={2026}
}

@inproceedings{spatialforcing,
  title={Spatial forcing: Implicit spatial representation alignment for vision-language-action model},
  author={Li, Fuhao and Song, Wenxuan and Zhao, Han and Wang, Jingbo and Ding, Pengxiang and Wang, Donglin and Zeng, Long and Li, Haoang},
  booktitle={International Conference on Learning Representations},
  volume={2026},
  pages={132324--132345},
  year={2026}
}

@article{xwam,
  title={Unified 4d world action modeling from video priors with asynchronous denoising},
  author={Guo, Jun and Li, Qiwei and Li, Peiyan and Chen, Zilong and Sun, Nan and Su, Yifei and Wang, Heyun and Zhang, Yuan and Li, Xinghang and Liu, Huaping},
  journal={arXiv preprint arXiv:2604.26694},
  year={2026}
}

@article{xiaomi1,
  title={Xiaomi-Robotics-1: Scaling Vision-Language-Action Models with over 100K Hours of Real-World Trajectories},
  author={Team, Xiaomi Robotics and Guo, Jun and Jin, Piaopiao and Li, Jason and Li, Peiyan and Li, Yingyan and Liu, Futeng and Peng, Wanli and Qin, Optimus and Su, Yifei and others},
  journal={arXiv preprint arXiv:2607.15330},
  year={2026}
}

@article{abotm0,
  title={Abot-m0: Vla foundation model for robotic manipulation with action manifold learning},
  author={Yang, Yandan and Zeng, Shuang and Lin, Tong and Chang, Xinyuan and Qi, Dekang and Xiao, Junjin and Liu, Haoyun and Chen, Ronghan and Chen, Yuzhi and Huo, Dongjie and others},
  journal={arXiv preprint arXiv:2602.11236},
  year={2026}
}

@article{starvla,
  title={StarVLA: A Lego-like Codebase for Vision-Language-Action Model Developing},
  author={Community, StarVLA},
  journal={arXiv preprint arXiv:2604.05014},
  year={2026}
}

@article{galaxea05,
  title={G0. 5: One autoregressive stream for robot reasoning and action},
  author={Liu, Yicheng and Dong, Zibin and Ye, Baijun and Yuan, Tianyuan and Jiang, Tao and Yang, Anqi and Cao, Shicheng and Liu, Haonan and Sun, Yue and Guo, Zihan and others},
  journal={arXiv preprint arXiv:2608.11739},
  year={2026}
}

@article{gigaworldpolicy,
  title={GigaWorld-Policy: An Efficient Action-Centered World--Action Model},
  author={Ye, Angen and Wang, Boyuan and Ni, Chaojun and Huang, Guan and Zhao, Guosheng and Li, Hao and Li, Hengtao and Li, Jie and Lv, Jindi and Liu, Jingyu and others},
  journal={arXiv preprint arXiv:2603.17240},
  year={2026}
}

@inproceedings{rdt,
  title={Rdt-1b: a diffusion foundation model for bimanual manipulation},
  author={Liu, Songming and Wu, Lingxuan and Li, Bangguo and Tan, Hengkai and Chen, Huayu and Wang, Zhengyi and Xu, Ke and Su, Hang and Zhu, Jun},
  booktitle={International Conference on Learning Representations},
  volume={2025},
  pages={29982--30009},
  year={2025}
}

@article{openvlaoft,
  title={Fine-tuning vision-language-action models: Optimizing speed and success},
  author={Kim, Moo Jin and Finn, Chelsea and Liang, Percy},
  journal={arXiv preprint arXiv:2502.19645},
  year={2025}
}

@misc{dm05,
    title  = {{DM0.5}: An Open-World Foundation Model for General-Purpose Embodied Intelligence},
    author = {{Dexmal Team}},
    month  = {July},
    year   = {2026},
    url    = {https://www.dexmal.com/blog/dm0.5/index_en.html}
}

@article{robodojo,
  title={RoboDojo: A unified sim-and-real benchmark for comprehensive evaluation of generalist robot manipulation policies},
  author={Chen, Tianxing and Chen, Yue and Li, Zixuan and Tang, Junyuan and Su, Kailun and Lu, Haoran and Wan, Weijie and Chen, Baijun and Liu, Songling and Yan, Haowen and others},
  journal={arXiv preprint arXiv:2607.04434},
  year={2026}
}

@article{libero,
  title={Libero: Benchmarking knowledge transfer for lifelong robot learning},
  author={Liu, Bo and Zhu, Yifeng and Gao, Chongkai and Feng, Yihao and Liu, Qiang and Zhu, Yuke and Stone, Peter},
  journal={Advances in Neural Information Processing Systems},
  volume={36},
  pages={44776--44791},
  year={2023}
}

@inproceedings{libero_plus,
  title={Libero-plus: A progressive robustness benchmark for visual-language-action models},
  author={Fei, Senyu and Wang, Siyin and Shi, Junhao and Dai, Zihao and Cai, Jikun and Qian, Pengfang and Ji, Li and He, Xinzhe and Zhang, Shiduo and Fei, Zhaoye and others},
  booktitle={Proceedings of the IEEE/CVF Conference on Computer Vision and Pattern Recognition},
  pages={38574--38583},
  year={2026}
}

@inproceedings{blip2,
  title={Blip-2: Bootstrapping language-image pre-training with frozen image encoders and large language models},
  author={Li, Junnan and Li, Dongxu and Savarese, Silvio and Hoi, Steven},
  booktitle={International conference on machine learning},
  pages={19730--19742},
  year={2023},
  organization={PmLR}
}

@article{llava,
  title={Visual instruction tuning},
  author={Liu, Haotian and Li, Chunyuan and Wu, Qingyang and Lee, Yong Jae},
  journal={Advances in neural information processing systems},
  volume={36},
  pages={34892--34916},
  year={2023}
}

@article{rt1,
  title={Rt-1: Robotics transformer for real-world control at scale},
  author={Brohan, Anthony and Brown, Noah and Carbajal, Justice and Chebotar, Yevgen and Dabis, Joseph and Finn, Chelsea and Gopalakrishnan, Keerthana and Hausman, Karol and Herzog, Alex and Hsu, Jasmine and others},
  journal={arXiv preprint arXiv:2212.06817},
  year={2022}
}

@article{rt2,
  title={Rt-2: Vision-language-action models transfer web knowledge to robotic control},
  author={Brohan, Anthony and Brown, Noah and Carbajal, Justice and Chebotar, Yevgen and Chen, Xi and Choromanski, Krzysztof and Ding, Tianli and Driess, Danny and Dubey, Avinava and Finn, Chelsea and others},
  journal={arXiv preprint arXiv:2307.15818},
  year={2023}
}

@article{openvla,
  title={Openvla: An open-source vision-language-action model},
  author={Kim, Moo Jin and Pertsch, Karl and Karamcheti, Siddharth and Xiao, Ted and Balakrishna, Ashwin and Nair, Suraj and Rafailov, Rafael and Foster, Ethan and Lam, Grace and Sanketi, Pannag and others},
  journal={arXiv preprint arXiv:2406.09246},
  year={2024}
}

@article{pi0,
  title = {{$\pi_0$: A Vision-Language-Action Flow Model for General Robot Control}},
  author = {Black, Kevin and Brown, Noah and Driess, Danny
            and Esmail, Adnan and Equi, Michael and Finn, Chelsea
            and Fusai, Niccolo and Groom, Lachy and Hausman, Karol
            and Ichter, Brian and others},
  journal = {arXiv preprint arXiv:2410.24164},
  year = {2024}
}

@article{dreamzero,
  title={World action models are zero-shot policies},
  author={Ye, Seonghyeon and Ge, Yunhao and Zheng, Kaiyuan and Gao, Shenyuan and Yu, Sihyun and Kurian, George and Indupuru, Suneel and Tan, You Liang and Zhu, Chuning and Xiang, Jiannan and others},
  journal={arXiv preprint arXiv:2602.15922},
  year={2026}
}

@article{lawam,
  title={Lawam: Latent world action models for efficient dynamics-aware robot policies},
  author={Chen, Jialei and Wang, Kai and Chen, Kang and Chen, Shuaihang and Gao, Feng and Tang, Wenhao and Li, Zhiyuan and Liu, Weilin and Yao, Zhuyu and Li, Boxun and others},
  journal={arXiv preprint arXiv:2606.15768},
  year={2026}
}

@article{bagelvla,
  title={Bagelvla: Enhancing long-horizon manipulation via interleaved vision-language-action generation},
  author={Hu, Yucheng and Zhang, Jianke and Luo, Yuanfei and Guo, Yanjiang and Chen, Xiaoyu and Sun, Xinshu and Feng, Kun and Lu, Qingzhou and Chen, Sheng and Zhang, Yangang and others},
  journal={arXiv preprint arXiv:2602.09849},
  year={2026}
}

@article{cosmos3,
  title={Cosmos 3: Omnimodal world models for physical ai},
  author={Agarwal, Niket and Ali, Arslan and Allen, Jon and Antolini, Martin and Aubame, Adeline and Azzolini, Alisson and Bai, Junjie and Bala, Maciej and Balaji, Yogesh and Bapst, Josh and others},
  journal={arXiv preprint arXiv:2606.02800},
  year={2026}
}

@article{lance,
  title={Lance: Unified multimodal modeling by multi-task synergy},
  author={Fu, Fengyi and Huang, Mengqi and Wu, Shaojin and Jiang, Yunsheng and Huo, Yufei and Li, Hao and Song, Yinghang and Ding, Fei and Guo, Jianzhu and He, Qian and others},
  journal={arXiv preprint arXiv:2605.18678},
  year={2026}
}

@article{wu2026foundation,
  title={From Foundation to Application: Improving VLA Models in Practice},
  author={Wu, Wei and Wang, Fangjing and Lu, Fan and Sun, He and Liu, Shi and Wang, Yunnan and Yan, Yibin and Wang, Yong and Ma, Shuailei and Wang, Xinyang and others},
  journal={arXiv preprint arXiv:2607.06403},
  year={2026}
}

@article{moge2,
  title={Moge-2: Accurate monocular geometry with metric scale and sharp details},
  author={Wang, Ruicheng and Xu, Sicheng and Dong, Yue and Deng, Yu and Xiang, Jianfeng and Lv, Zelong and Sun, Guangzhong and Tong, Xin and Yang, Jiaolong},
  journal={Advances in Neural Information Processing Systems},
  volume={38},
  pages={35928--35959},
  year={2026}
}

@inproceedings{waft,
  title={Waft: Warping-alone field transforms for optical flow},
  author={Wang, Yihan and Deng, Jia},
  booktitle={International Conference on Learning Representations},
  volume={2026},
  pages={157389--157402},
  year={2026}
}

@inproceedings{genception,
  author    = {Letian Wang and Chuhan Zhang and Rishabh Kabra and Jasper Uijlings and Steven Waslander and Andrew Zisserman and Joao Carreira and Kaiming He and Misha Andriluka and Eduard Gabriel Bazavan and Andrei Zanfir and Cristian Sminchisescu},
  title     = {Video Generation Models are General-Purpose Vision Learners},
  booktitle = {European Conference on Computer Vision (ECCV)},
  year      = {2026},
  url       = {https://genception.github.io/},
}

@inproceedings{flow_matching,
  author    = {Yaron Lipman and Ricky T. Q. Chen and Heli Ben-Hamu and Maximilian Nickel and Matthew Le},
  title     = {Flow Matching for Generative Modeling},
  booktitle = {The Eleventh International Conference on Learning Representations},
  year      = {2023},
  url       = {https://openreview.net/forum?id=PqvMRDCJT9t},
}

@article{octo,
  title={Octo: An open-source generalist robot policy},
  author={Team, Octo Model and Ghosh, Dibya and Walke, Homer and Pertsch, Karl and Black, Kevin and Mees, Oier and Dasari, Sudeep and Hejna, Joey and Kreiman, Tobias and Xu, Charles and others},
  journal={arXiv preprint arXiv:2405.12213},
  year={2024}
}

@article{fast,
  title={Fast: Efficient action tokenization for vision-language-action models},
  author={Pertsch, Karl and Stachowicz, Kyle and Ichter, Brian and Driess, Danny and Nair, Suraj and Vuong, Quan and Mees, Oier and Finn, Chelsea and Levine, Sergey},
  journal={arXiv preprint arXiv:2501.09747},
  year={2025}
}

@article{unipi,
  title={Learning universal policies via text-guided video generation},
  author={Du, Yilun and Yang, Sherry and Dai, Bo and Dai, Hanjun and Nachum, Ofir and Tenenbaum, Josh and Schuurmans, Dale and Abbeel, Pieter},
  journal={Advances in neural information processing systems},
  volume={36},
  pages={9156--9172},
  year={2023}
}

@inproceedings{gr1,
  title={Unleashing large-scale video generative pre-training for visual robot manipulation},
  author={Wu, Hongtao and Jing, Ya and Cheang, Chilam and Chen, Guangzeng and Xu, Jiafeng and Li, Xinghang and Liu, Minghuan and Li, Hang and Kong, Tao},
  booktitle={International Conference on Learning Representations},
  volume={2024},
  pages={10641--10662},
  year={2024}
}

@inproceedings{showo,
  title={Show-o: One single transformer to unify multimodal understanding and generation},
  author={Xie, Jinheng and Mao, Weijia and Bai, Zechen and Zhang, David Junhao and Wang, Weihao and Lin, Kevin Qinghong and Gu, Yuchao and Chen, Zhijie and Yang, Zhenheng and Shou, Mike Zheng},
  booktitle={International Conference on Learning Representations},
  volume={2025},
  pages={28240--28264},
  year={2025}
}

@article{bagel,
  title={Emerging properties in unified multimodal pretraining},
  author={Deng, Chaorui and Zhu, Deyao and Li, Kunchang and Gou, Chenhui and Li, Feng and Wang, Zeyu and Zhong, Shu and Yu, Weihao and Nie, Xiaonan and Song, Ziang and others},
  journal={arXiv preprint arXiv:2505.14683},
  year={2025}
}

@article{pi05,
  author  = {{Physical Intelligence} and Kevin Black and Noah Brown and James Darpinian and Karan Dhabalia and Danny Driess and Adnan Esmail and Michael Equi and Chelsea Finn and others},
  title   = {{$\pi_{0.5}$}: A Vision-Language-Action Model with Open-World Generalization},
  journal = {arXiv preprint arXiv:2504.16054},
  year    = {2025},
  url     = {https://arxiv.org/abs/2504.16054},
}

@article{fastwam,
  title={Fast-wam: Do world action models need test-time future imagination?},
  author={Yuan, Tianyuan and Dong, Zibin and Liu, Yicheng and Zhao, Hang},
  journal={arXiv preprint arXiv:2603.16666},
  year={2026}
}

@article{uam,
  author  = {Jianke Zhang and Yuanfei Luo and Yucheng Hu and Xiaoyu Chen and Yanjiang Guo and Ziyang Liu and Hongbin Xu and Tian Lan and Jianyu Chen},
  title   = {{UAM}: A Dual-Stream Perspective on Forgetting in {VLA} Training},
  journal = {arXiv preprint arXiv:2605.15735},
  year    = {2026},
  url     = {https://arxiv.org/abs/2605.15735},
}

@inproceedings{motus,
  author    = {Hongzhe Bi and Hengkai Tan and Shenghao Xie and Zeyuan Wang and others},
  title     = {{Motus}: A Unified Latent Action World Model},
  booktitle = {Proceedings of the IEEE/CVF Conference on Computer Vision and Pattern Recognition (CVPR)},
  year      = {2026},
  url       = {https://arxiv.org/abs/2512.13030},
}

@article{motubrain,
  title={Motubrain: An advanced world action model for robot control},
  author={Team, MotuBrain and Xiang, Chendong and Bao, Fan and Liu, Haitian and Tan, Hengkai and Bi, Hongzhe and Li, James and Liu, Jiabao and Pang, Jingrui and Jing, Kiro and others},
  journal={arXiv preprint arXiv:2604.27792},
  year={2026}
}

@article{imagewam,
  title={ImageWAM: Do World Action Models Really Need Video Generation, or Just Image Editing?},
  author={Zhang, Yuyang and Zhang, Wenyao and Qi, Zekun and Zhang, He and Lin, Haitao and Zhang, Jingbo and Mu, Yao and Yang, Xiaokang and Zeng, Wenjun and Jin, Xin},
  journal={arXiv preprint arXiv:2606.19531},
  year={2026}
}

@article{beingh07,
  title = {{Being-H0.7}: A Latent World-Action Model from Egocentric Videos},
  author = {Hao Luo and Wanpeng Zhang and Yicheng Feng and Sipeng Zheng
            and Haiweng Xu and Chaoyi Xu and Ziheng Xi and Yuhui Fu
            and Zongqing Lu},
  journal = {arXiv preprint arXiv:2605.00078},
  year = {2026},
  url = {https://arxiv.org/abs/2605.00078}
}

@inproceedings{cosmospolicy,
  title = {{Cosmos Policy}: Fine-Tuning Video Models for Visuomotor Control and Planning},
  author = {Moo Jin Kim and Yihuai Gao and Tsung-Yi Lin and Yen-Chen Lin
            and Yunhao Ge and Grace Lam and Percy Liang and Shuran Song
            and Ming-Yu Liu and Chelsea Finn and Jinwei Gu},
  booktitle = {International Conference on Learning Representations},
  year = {2026},
  url = {https://arxiv.org/abs/2601.16163}
}

@article{yu2023inpaint,
  title={Inpaint anything: Segment anything meets image inpainting},
  author={Yu, Tao and Feng, Runseng and Feng, Ruoyu and Liu, Jinming and Jin, Xin and Zeng, Wenjun and Chen, Zhibo},
  journal={arXiv preprint arXiv:2304.06790},
  year={2023}
}

@article{lin2025depth,
  title={Depth anything 3: Recovering the visual space from any views},
  author={Lin, Haotong and Chen, Sili and Liew, Junhao and Chen, Donny Y and Li, Zhenyu and Shi, Guang and Feng, Jiashi and Kang, Bingyi},
  journal={arXiv preprint arXiv:2511.10647},
  year={2025}
}

@inproceedings{carion2026sam,
  title={Sam 3: Segment anything with concepts},
  author={Carion, Nicolas and Gustafson, Laura and Hu, Yuan-Ting and Debnath, Shoubhik and Hu, Ronghang and Suris Coll-Vinent, Didac and Ryali, Chaitanya and Alwala, Kalyan Vasudev and Khedr, Haitham and Huang, Andrew and others},
  booktitle={International conference on learning representations},
  volume={2026},
  pages={138846--138923},
  year={2026}
}

@inproceedings{sam3dteam2025sam3d3dfyimages,
  title={Sam 3d: 3dfy anything in images},
  author={Chen, Xingyu and Chu, Fu-Jen and Gleize, Pierre and Liang, Kevin J and Sax, Alexander and Tang, Hao and Wang, Weiyao and Guo, Michelle and Hardin, Thibaut and Li, Xiang and others},
  booktitle={Proceedings of the IEEE/CVF Conference on Computer Vision and Pattern Recognition},
  pages={7220--7232},
  year={2026}
}

@article{pan2025spiderscalablephysicsinformeddexterous,
  title={Spider: Scalable physics-informed dexterous retargeting},
  author={Pan, Chaoyi and Wang, Changhao and Qi, Haozhi and Liu, Zixi and Bharadhwaj, Homanga and Sharma, Akash and Wu, Tingfan and Shi, Guanya and Malik, Jitendra and Hogan, Francois},
  journal={arXiv preprint arXiv:2511.09484},
  year={2025}
}

@article{lai2026a,
  title   = {CoWTracker: Tracking by Warping instead of Correlation},
  author  = {Lai, Zihang and Insafutdinov, Eldar and Sucar, Edgar and Vedaldi, Andrea},
  journal = {arXiv preprint arXiv:2602.04877},
  year    = {2026},
}

@inproceedings{wang2025vggt,
  title={VGGT: Visual Geometry Grounded Transformer},
  author={Wang, Jianyuan and Chen, Minghao and Karaev, Nikita and Vedaldi, Andrea and Rupprecht, Christian and Novotny, David},
  booktitle={Proceedings of the IEEE/CVF Conference on Computer Vision and Pattern Recognition},
  year={2025}
}

@inproceedings{zhou2023propainter,
   title={{ProPainter}: Improving Propagation and Transformer for Video Inpainting},
   author={Zhou, Shangchen and Li, Chongyi and Chan, Kelvin C.K and Loy, Chen Change},
   booktitle={Proceedings of IEEE International Conference on Computer Vision (ICCV)},
   year={2023}
}

@InProceedings{foundationposewen2024,
author        = {Bowen Wen, Wei Yang, Jan Kautz and Stan Birchfield},
title         = {{FoundationPose}: Unified 6D Pose Estimation and Tracking of Novel Objects},
booktitle     = {CVPR},
year          = {2024},
}

@inproceedings{todorov2012mujoco,
  title={MuJoCo: A physics engine for model-based control},
  author={Todorov, Emanuel and Erez, Tom and Tassa, Yuval},
  booktitle={2012 IEEE/RSJ International Conference on Intelligent Robots and Systems},
  pages={5026--5033},
  year={2012},
  organization={IEEE},
  doi={10.1109/IROS.2012.6386109}
}

@article{punamiya2026egoverse,
  title={Egoverse: An egocentric human dataset for robot learning from around the world},
  author={Punamiya, Ryan and Kareer, Simar and Liu, Zeyi and Citron, Josh and Qiu, Ri-Zhao and Cai, Xiongyi and Gavryushin, Alexey and Chen, Jiaqi and Liconti, Davide and Zhu, Lawrence Y and others},
  journal={arXiv preprint arXiv:2604.07607},
  year={2026}
}

@article{li2026egolive,
  title={Egolive: A large-scale egocentric dataset from real-world human tasks},
  author={Li, Yihang and Wei, Xuelong and Luo, Jingzhou and Xiao, Yingjing and Bai, Yibo and Zhou, Guangyuan and Zou, Teng and Gui, Chenguang and Wen, Jiajun and Zhang, He and others},
  journal={arXiv preprint arXiv:2604.23570},
  year={2026}
}

@article{zhang2026joyai,
  title={Joyai-ra 0.1: A foundation model for robotic autonomy},
  author={Zhang, Tianle and Yuan, Zhihao and Chi, Dafeng and Liu, Peidong and Li, Dongwei and Hu, Kejun and Zhang, Likui and Nie, Junnan and Wei, Ziming and Chen, Zengjue and others},
  journal={arXiv preprint arXiv:2604.20100},
  year={2026}
}

@inproceedings{hoque2026egodex,
  title={Egodex: Learning dexterous manipulation from large-scale egocentric video},
  author={Hoque, Ryan and Huang, Peide and Yoon, David and Zhang, Jian and others},
  booktitle={International Conference on Learning Representations},
  volume={2026},
  pages={4218--4237},
  year={2026}
}

@inproceedings{grauman2024ego,
  title={Ego-exo4d: Understanding skilled human activity from first-and third-person perspectives},
  author={Grauman, Kristen and Westbury, Andrew and Torresani, Lorenzo and Kitani, Kris and Malik, Jitendra and Afouras, Triantafyllos and Ashutosh, Kumar and Baiyya, Vijay and Bansal, Siddhant and Boote, Bikram and others},
  booktitle={Proceedings of the IEEE/CVF conference on computer vision and pattern recognition},
  pages={19383--19400},
  year={2024}
}

@inproceedings{liu2022hoi4d,
  title={Hoi4d: A 4d egocentric dataset for category-level human-object interaction},
  author={Liu, Yunze and Liu, Yun and Jiang, Che and Lyu, Kangbo and Wan, Weikang and Shen, Hao and Liang, Boqiang and Fu, Zhoujie and Wang, He and Yi, Li},
  booktitle={2022 IEEE/CVF Conference on Computer Vision and Pattern Recognition (CVPR)},
  pages={20981--20990},
  year={2022},
  organization={IEEE}
}

@inproceedings{banerjee2025hot3d,
  title={Hot3d: Hand and object tracking in 3d from egocentric multi-view videos},
  author={Banerjee, Prithviraj and Shkodrani, Sindi and Moulon, Pierre and Hampali, Shreyas and Han, Shangchen and Zhang, Fan and Zhang, Linguang and Fountain, Jade and Miller, Edward and Basol, Selen and others},
  booktitle={2025 IEEE/CVF Conference on Computer Vision and Pattern Recognition (CVPR)},
  pages={7061--7071},
  year={2025},
  organization={IEEE}
}

@article{yuan2026qwen,
  title={Qwen-robotmanip technical report: Alignment unlocks scale for robotic manipulation foundation models},
  author={Yuan, Haoqi and Liang, Zhixuan and Chen, Anzhe and Wang, Ye and Li, Haoyang and Lin, Pei and Huang, Yiyang and Lei, Zixing and Zhang, Tong and Zhang, Jiazhao and others},
  journal={arXiv preprint arXiv:2606.17846},
  year={2026}
}

@article{han2026video2sim2real,
  title={Video2Sim2Real: Full-Stack Autonomous Dexterous Skill Acquisition from a Single Human Video},
  author={Han, Yunhai and Qiu, Jianuo and Bai, Linhao and Xiao, Ziyu and Zeng, Zihang and Liu, Yangcen and Yang, Zhaodong and Jain, Shalin and Ma, Wenrui and Fu, Jiaqi and others},
  journal={arXiv preprint arXiv:2606.08828},
  year={2026}
}

@article{liu2026egoengine,
  title={EgoEngine: From Egocentric Human Videos to High-Fidelity Dexterous Robot Demonstrations},
  author={Liu, Yangcen and Cheng, Shuo and Yin, Xinchen and Shin, Woo Chul and Cueva, Alfred and Yang, Yiran and Chen, Zhenyang and Zhang, Chuye and Xu, Danfei},
  journal={arXiv preprint arXiv:2606.12604},
  year={2026}
}

@article{wang2026ego2robot,
  title={Ego2Robot: Scalable Robot Data Synthesis from Egocentric Human Data},
  author={Wang, Ye and Lin, Pei and Chen, Xiong-Hui and Yuan, Haoqi and Liang, Zhixuan and Huang, Yiyang and Chen, Anzhe and Lei, Zixing and Zhang, Jie and Zhang, Tao and others},
  journal={arXiv preprint arXiv:2608.02580},
  year={2026}
}

@article{yang2026handedit,
  title={HandEdit: A Unified Benchmark for Egocentric Human-to-Robot Dexterous Hand Image Editing},
  author={Yang, Zhenjie and Jiao, Xingyu and Zhong, Guopeng and Yang, Shuzhe and Che, Shi and Wu, Chao and Jiang, Chenyu and Zhang, Dongjie and Zhang, Yideng and Zhang, Zheng and others},
  journal={arXiv preprint arXiv:2608.12122},
  year={2026}
}

@article{schulman2017proximal,
  title={Proximal policy optimization algorithms},
  author={Schulman, John and Wolski, Filip and Dhariwal, Prafulla and Radford, Alec and Klimov, Oleg},
  journal={arXiv preprint arXiv:1707.06347},
  year={2017}
}

@article{yu2026affordancevla,
  title={AffordanceVLA: A Vision-Language-Action Model Empowering Action Generation through Affordance-Aware Understanding},
  author={Yu, Qize and You, Jiadi and Wang, Yuran and Liang, Jiaqi and Ping, Bowen and Tian, Yang and Chen, Yue and Cai, Minghong and Gong, Zeying and Wu, Ruihai and others},
  journal={arXiv preprint arXiv:2606.06155},
  year={2026}
}

@misc{qwen36_35b_a3b,
    title = {{Qwen3.6-35B-A3B}: Agentic Coding Power, Now Open to All},
    url = {https://qwen.ai/blog?id=qwen3.6-35b-a3b},
    author = {{Qwen Team}},
    month = {April},
    year = {2026}
}

@misc{openai2026gpt55,
  author       = {{OpenAI}},
  title        = {Introducing GPT-5.5},
  year         = {2026},
  month        = apr,
  day          = {23},
  howpublished = {\url{https://openai.com/zh-Hans-CN/index/introducing-gpt-5-5/}},
  note         = {Accessed: 2026-09-10}
}

@misc{bai2025qwen25vltechnicalreport,
  title = {{Qwen2.5-VL} Technical Report},
  author = {Shuai Bai and Keqin Chen and Xuejing Liu and Jialin Wang and Wenbin Ge and Sibo Song and Kai Dang and Peng Wang and Shijie Wang and Jun Tang and Humen Zhong and Yuanzhi Zhu and Mingkun Yang and Zhaohai Li and Jianqiang Wan and Pengfei Wang and Wei Ding and Zheren Fu and Yiheng Xu and Jiabo Ye and Xi Zhang and Tianbao Xie and Zesen Cheng and Hang Zhang and Zhibo Yang and Haiyang Xu and Junyang Lin},
  year = {2025},
  eprint = {2502.13923},
  archivePrefix = {arXiv},
  primaryClass = {cs.CV},
  url = {https://arxiv.org/abs/2502.13923},
}

@article{bu2025agibot,
  title={AgiBot world colosseo: A large-scale manipulation platform for scalable and intelligent embodied systems},
  author={Bu, Qingwen and Cai, Jisong and Chen, Li and Cui, Xiuqi and Ding, Yan and Feng, Siyuan and Gao, Shenyuan and others},
  journal={arXiv preprint arXiv:2503.06669},
  year={2025}
}

@article{hou2026robomind20multimodalbimanual,
  title={Robomind 2.0: A multimodal, bimanual mobile manipulation dataset for generalizable embodied intelligence},
  author={Hou, Chengkai and Wu, Kun and Liu, Jiaming and Che, Zhengping and Wu, Di and Liao, Fei and Li, Guangrun and He, Jingyang and Feng, Qiuxuan and Jin, Zhao and others},
  journal={arXiv preprint arXiv:2512.24653},
  year={2025}
}

@misc{wu2026robocoinopensourcedbimanualrobotic,
      title={RoboCOIN: An Open-Sourced Bimanual Robotic Data Collection for Integrated Manipulation}, 
      author={Shihan Wu and others},
      year={2026},
      eprint={2511.17441},
      archivePrefix={arXiv},
      primaryClass={cs.RO},
      url={https://arxiv.org/abs/2511.17441},
}

@article{galaxea2025,
  title={Galaxea G0: Open-World Dataset and Dual-System VLA Model},
  author={Galaxea Team},
  journal={arXiv preprint arXiv:2509.00576},
  year={2025}
}

@misc{shi2026diversityneedscalablerobotic,
      title={Is Diversity All You Need for Scalable Robotic Manipulation?}, 
      author={Modi Shi and Li Chen and Jin Chen and Yuxiang Lu and Chiming Liu and Guanghui Ren and Ping Luo and Di Huang and Maoqing Yao and Hongyang Li},
      year={2026},
      eprint={2507.06219},
      archivePrefix={arXiv},
      primaryClass={cs.RO},
      url={https://arxiv.org/abs/2507.06219}, 
}

@article{intelligence2026pi07steerablegeneralistrobotic,
  title={${\pi}_{0.7}$: a Steerable Generalist Robotic Foundation Model with Emergent Capabilities},
  author={Intelligence, Physical and Ai, Bo and Amin, Ali and Aniceto, Raichelle and Balakrishna, Ashwin and Balke, Greg and Black, Kevin and Bokinsky, George and Cao, Shihao and Charbonnier, Thomas and others},
  journal={arXiv preprint arXiv:2604.15483},
  year={2026}
}

@misc{cadene2024lerobot,
    author = {Cadene, Remi and Alibert, Simon and Soare, Alexander and Gallouedec, Quentin and Zouitine, Adil and Palma, Steven and Kooijmans, Pepijn and Aractingi, Michel and Shukor, Mustafa and Aubakirova, Dana and Russi, Martino and Capuano, Francesco and Pascal, Caroline and Choghari, Jade and Meftah, Khalil and Ellerbach, Maxime and Moss, Jess and Wolf, Thomas},
    title = {LeRobot: State-of-the-art Machine Learning for Real-World Robotics in Pytorch},
    howpublished = "\url{https://github.com/huggingface/lerobot}",
    year = {2024}
}

@article{li2026causal,
  title={Causal world modeling for robot control},
  author={Li, Lin and Zhang, Qihang and Luo, Yiming and Yang, Shuai and Wang, Ruilin and Han, Fei and Yu, Mingrui and Gao, Zelin and Xue, Nan and Zhu, Xing and others},
  journal={arXiv preprint arXiv:2601.21998},
  year={2026}
}

\appendix

\end{document}